\documentclass{article}

\usepackage{microtype}
\usepackage{graphicx}
\usepackage{booktabs} 
\usepackage{caption}
\usepackage{subcaption}
\usepackage{bbding}
\usepackage{multirow}
\usepackage{hyperref}

\usepackage[accepted]{icml2023}

\usepackage{amsmath}
\usepackage{amssymb}
\usepackage{mathtools}
\usepackage{amsthm}
\usepackage{comment}
\usepackage{wrapfig}
\usepackage[capitalize,noabbrev]{cleveref}

\theoremstyle{plain}

\theoremstyle{definition}

\theoremstyle{remark}

\usepackage[textsize=tiny]{todonotes}

\usepackage{tikz}
\usepackage{pgfplots}
\pgfplotsset{compat=1.16}
\usepgfplotslibrary{external} 
\icmltitlerunning{Pythia: A Suite for Analyzing Large Language Models}

\begin{document}

\twocolumn[
\icmltitle{\textit{Pythia}: A Suite for Analyzing Large Language Models\\ Across Training and Scaling}



\icmlsetsymbol{equal}{*}

\begin{icmlauthorlist}
\icmlauthor{Stella Biderman}{equal,eai,booz}
\icmlauthor{Hailey Schoelkopf}{equal,eai,yale}
\icmlauthor{Quentin Anthony}{eai}
\icmlauthor{Herbie Bradley}{eai,cambridge}
\icmlauthor{Kyle O'Brien}{eai}
\icmlauthor{Eric Hallahan}{eai}
\icmlauthor{Mohammad Aflah Khan}{iiit}
\icmlauthor{Shivanshu Purohit}{sai,eai}
\icmlauthor{USVSN Sai Prashanth}{eai}
\icmlauthor{Edward Raff}{booz}
\icmlauthor{Aviya Skowron}{eai}
\icmlauthor{Lintang Sutawika}{eai,dai}
\icmlauthor{Oskar van der Wal}{uoa}
\end{icmlauthorlist}

\icmlaffiliation{eai}{EleutherAI}
\icmlaffiliation{booz}{Booz Allen Hamilton, McLean, USA}
\icmlaffiliation{yale}{Yale University, New Haven, USA}
\icmlaffiliation{cambridge}{University of Cambridge, UK}
\icmlaffiliation{iiit}{Indraprastha Institute of Information Technology Delhi, India}
\icmlaffiliation{sai}{Stability AI}
\icmlaffiliation{uoa}{Institute for Logic, Language and Computation, University of Amsterdam, Netherlands}
\icmlaffiliation{dai}{Datasaur.ai, USA}
\icmlcorrespondingauthor{Stella Biderman}{stella@eleuther.ai}
\icmlcorrespondingauthor{Hailey Schoelkopf}{hailey@eleuther.ai}

\icmlkeywords{Machine Learning, ICML}

\vskip 0.3in
]



\printAffiliationsAndNotice{\icmlEqualContribution} 

\begin{abstract}
How do large language models (LLMs) develop and evolve over the course of training? How do these patterns change as models scale? To answer these questions, we introduce \textit{Pythia}, a suite of 16 LLMs all trained on public data seen in the exact same order and ranging in size from 70M to 12B parameters. We provide public access to 154 checkpoints for each one of the 16 models, alongside tools to download and reconstruct their exact training dataloaders for further study. We intend \textit{Pythia} to facilitate research in many areas, and we present several case studies including novel results in memorization, term frequency effects on few-shot performance, and reducing gender bias. We demonstrate that this highly controlled setup can be used to yield novel insights toward LLMs and their training dynamics. Trained models, analysis code, training code, and training data can be found at \url{https://github.com/EleutherAI/pythia}.
\end{abstract}

\section{Introduction}\label{sec:introduction}
Over the past several years, large transformer models have established themselves as the premier methodology for generative tasks in natural language processing \citep{brown2020language,sanh2021multitask,chowdhery2022palm}. Beyond NLP, transformers have also made big splashes as generative models in areas as diverse as text-to-image synthesis \citep{ramesh2022hierarchical,crowson2022vqgan,rombach2022high}, protein modeling \citep{jumper2021highly,ahdritz2022openfold}, and computer programming \citep{chen2021evaluating,xu2022systematic,fried2022incoder}. Despite these successes, very little is known about how and why these models are so successful.

Critical to understanding the functioning of transformers is better understanding how these models behave along two axes: training and scaling. It is well established that there are regular and predictable patterns in the behavior of trained language models as they scale \citep{kaplan2020scaling,henighan2020scaling,hernandez2021scaling,mikami2021scaling,pu2021scaling,sharma2020neural,ghorbani2021scaling}, but prior work connecting these ``Scaling Laws'' to the learning dynamics of language models is minimal. One of the driving reasons for this gap in research is a lack of access to appropriate model suites to test theories: although there are more publicly available LLMs than ever, they do not meet common requirements for researchers, as discussed in \cref{sec:pythia} of this paper.  Of the research along
these lines that does exist \citep{mcgrath2021acquisition,tirumala2022memorization,xia2022training},  it is overwhelmingly done on non-public models or model checkpoints, further emphasizing the importance of having publicly available model suites
for scientific research.

\begin{table*}[ht]
    \centering
    \begin{tabular}{rrcccccc}\toprule
  Model Size & Non-Embedding Params & Layers & Model Dim & Heads & Learning Rate        & Equivalent Models\\\midrule
          70 M  &     18,915,328   &  6     &  512      &  8    & $10.0\times 10^{-4}$  & ---\\
         160 M  &     85,056,000   & 12     &  768      & 12    & $6.0\times 10^{-4}$  & GPT-Neo 125M, OPT-125M\\
         410 M  &    302,311,424   & 24     & 1024      & 16    & $3.0\times 10^{-4}$  & OPT-350M\\
         1.0 B  &    805,736,448   & 16     & 2048      &  8    & $3.0\times 10^{-4}$  & --- \\
         1.4 B  &  1,208,602,624   & 24     & 2048      & 16    & $2.0\times 10^{-4}$  & GPT-Neo 1.3B, OPT-1.3B\\
         2.8 B  &  2,517,652,480   & 32     & 2560      & 32    & $1.6\times 10^{-4}$  & GPT-Neo 2.7B, OPT-2.7B\\
         6.9 B  &  6,444,163,072   & 32     & 4096      & 32    & $1.2\times 10^{-4}$  & OPT-6.7B\\
          12 B  & 11,327,027,200   & 36     & 5120      & 40    & $1.2\times 10^{-4}$  & ---\\\bottomrule
	\end{tabular}
	\caption{Models in the Pythia suite and select hyperparameters. For a full list of hyper-parameters, see \Cref{app:hparam}. Models are named based on their total number of parameters, but for most analyses we recommend people use the number of non-embedding parameters as the measure of ``size.'' Models marked as ``equivalent'' have the same architecture and number of non-embedding parameters.}
	\label{table:interp}
\end{table*}

In this paper we introduce \textit{Pythia}, a suite of decoder-only autoregressive language models ranging from 70M to 12B parameters designed specifically to facilitate such scientific research. The \textit{Pythia} suite is the only publicly released suite of LLMs that satisfies three key properties:
\begin{enumerate}
    \item Models span several orders of magnitude of model scale.
    \item All models were trained on the same data in the same order.
    \item The data and intermediate checkpoints are publicly available for study.
\end{enumerate}
We train 8 model sizes each on both the Pile \cite{gao2020pile,biderman2022datasheet} and the Pile after deduplication, providing 2 copies of the suite which can be compared.

We use these key properties of \textit{Pythia} in order to study for the first time how properties like gender bias, memorization, and few-shot learning are affected by the precise training data processed and model scale. We intend the following experiments to be case studies demonstrating the experimental setups \textit{Pythia} enables, and to additionally provide directions for future work.

\paragraph{Mitigating Gender Bias} There is much work cataloging how language models reflect the biases encoded in their training data. However, while some work has explored finetuning's effects on bias in language models \citep{gira2022debiasing,kirtane2022efficient,choenni2021stepmothers}, or the relationship between the corpus statistics and the measured bias \citep{bordia2019identifying,wal2022birth}, researchers have generally lacked the tools to study the role of the training data on the learning dynamics of bias in large language models of different sizes. To demonstrate what is now possible with Pythia, we analyze whether deliberately modifying the frequency of gendered terms in the pretraining data of a language model can have an impact on its downstream behavior and biases. We leverage the known pretraining data and public training codebase of our model suite, and counterfactually retrain models such that the last 7\% and 21\% of model training has a majority of pronouns modified such that their grammatical gender is feminine rather than masculine. We demonstrate that such interventions are successful at reducing bias measures on a targeted benchmark, and propose counterfactual interventions and retrainability of portions of our models as a key tool for future study of the influence of training corpora on model behavior.

\paragraph{Memorization is a Poisson Point Process} Building on the extensive literature on memorization in large language models \citep{carlini2019secret,carlini2021extracting,hu2022membership}, we ask the following question: does the location of a particular sequence in the training dataset influence the likelihood of it being memorized? Leveraging \textit{Pythia}'s reproducible dataloader setup we answer this question in the negative, and furthermore find that a poisson point process is a very good model for the occurrence of memorized sequences over the course of training.

\paragraph{Emergence of the Impact of Pretraining Frequencies} Recent work has identified the frequency of specific facts within a corpus as an important factor in how likely a model is capable of applying that fact in response to a natural language question \citep{razeghi2022impact,elazar2022measuring,kandpal2022large,mallen2022parametric}. Existing work has been heavily dependent on the handful of models trained on public data, such as GPT-J \citep{gpt-j} and BLOOM \citep{scao2022bloom}, which lack frequent intermediate checkpoints, so none of these papers are able to look at the fine-grained evolution of this phenomenon over the course of training. To address this gap in the literature, we examine how the role of pretraining term frequencies changes over the course of training. We find that a significant phase change occurs after 65,000 training steps (45\% through training): the models with 2.8 billion parameters or more start to exhibit a correlation between task accuracy and occurrence of task-relevant terms which is not present in prior checkpoints and which is largely absent from smaller models.

\section{The Pythia Suite}\label{sec:pythia}

Following the advice of \citet{birhane2021values}, in this section we seek to explicitly document our choices, rationales, and values in designing and implementing \textit{Pythia}. As our goal is to promote scientific research on large language models, we prioritize consistency in model design and controlling for as many potential sources of variation as possible, rather than trying to eke out the most performance from each model. For example, we use the parallel attention and feedforward approach for all models, as it is becoming widely used for the largest models, even though it is generally not recommended for models with less than 6B parameters. To our surprise, we find that despite making choices we expect to hurt performance at smaller scales, we find that our models perform the same as equi-parameter OPT models across all scales. We discuss areas where our results contradict widely accepted maxims for training LLMs in \Cref{sec:novel}.

\subsection{Requirements for a Scientific Suite of LLMs}

Pythia is envisioned as a suite for enabling and empowering scientific research on the capacities and limitations of large language models. After surveying the existing literature, we found no existing suite of models which satisfied all of the following conditions:

\paragraph{Public Access} Models are publicly released and are trained on publicly available data.

\paragraph{Training Provenance} Intermediate checkpoints are available for analysis, all models are trained with the same data ordering, and intermediate checkpoints can be linked with the exact data seen up to that checkpoint. Training procedure as well as model and training hyperparameters are well-documented.

\begin{table*}[htb]
\centering
\begin{tabular}{rccccccc}\toprule
                          & GPT-2        & GPT-3           & GPT-Neo         & OPT             & T5           & BLOOM           & Pythia (ours) \\ \midrule
Public Models             & \CircleSolid & \HalfCircleLeft & \CircleSolid    & \CircleSolid    & \CircleSolid & \CircleSolid    & \CircleSolid  \\
Public Data               &              &                 & \CircleSolid    &                 & \CircleSolid & \HalfCircleLeft & \CircleSolid  \\ \midrule
Known Training Order      &              &                 & \CircleSolid    &                 &              & \HalfCircleLeft & \CircleSolid  \\ 
Consistent Training Order &              &                 &                 & \CircleSolid    &              & \HalfCircleLeft & \CircleSolid  \\
Number of Checkpoints     & 1            & 1               & 30               & 2               & 1            & 8               & 154           \\\midrule
Smallest Model            & 124M         & Ada             & 125M            & 125M            & 60M          & 560M            & 70M           \\
Largest Model             & 1.5B         & DaVinci         & 20B             & 175B            & 11B          & 176B            & 12B           \\
Number of Models          & 4            & 4               & 6               & 9               & 5            & 5               & 8             \\\bottomrule
\end{tabular}
\caption{Commonly used model suites and how they rate according to our requirements. Further information can be found in \cref{app:suites}.}
\label{table:reqs}
\end{table*}

\paragraph{Consistency Across Scale} Model scaling sequences should have self-consistent design decisions that reasonably adhere to common practice for training state-of-the-art large models. Model sizes should cover a variety of scales across multiple orders of magnitude.

\Cref{table:reqs} provides our assessment of a number of popular language model suites along these criteria. We note that for ``number of checkpoints'' we go with the number of checkpoints by the model in the model suite with \textit{the fewest checkpoints}. While some model suites (e.g., GPT-Neo, OPT, BLOOM) have a subset that have more available, for most research purposes this is insufficient. This is exacerbated by the fact that typically smaller models are the ones with more checkpoints; the only model suite from the above list whose largest model has more checkpoints than smaller ones is GPT-Neo.

\subsection{Training Data}

We train our models on the Pile \citep{gao2020pile,biderman2022datasheet}, a curated collection of English language datasets for training large language models that is popular for training large autoregressive transformers. This dataset has three major benefits over its competitors: first, it is freely and publicly available; second, it reports a higher downstream performance \citep{le2022language} than popular crawl-based datasets C4 \citep{raffel2020exploring,dodge2021documenting} and OSCAR \citep{OSCAR}; and third, it has been widely used by state-of-the-art models including \mbox{GPT-J-6B} \citep{gpt-j}, \mbox{GPT-NeoX-20B} \citep{black2022gpt}, \mbox{Jurassic-1} \citep{lieber2021jurassic}\footnote{While the paper discusses the Pile at length, it does not explicitly state that Jurassic-1 was trained on the Pile. We originally discovered this fact by executing data extraction attacks on the API, and confirmed with private communication with the authors.}, Megatron-Turing NLG 530B \citep{Megatron-Turing}, OPT \citep{zhang2022opt}, and WuDao \citep{WuDao}. We use the tokenizer developed by \citet{black2022gpt}, which is a BPE tokenizer that is trained specifically on the Pile.

While we considered training on a multilingual corpus instead of a monolingual one, we ultimately opted against doing so for the following reasons:
\begin{enumerate}
    \item While we are confident that we are generally aware of the contents and quality of the Pile, we cannot say the same for multilingual datasets. Existing massive multilingual datasets can be of dubious quality \citep{caswell2020language,kreutzer2021quality} and we do not feel qualified to vet existing multilingual datasets well enough to determine issues that may arise due to using them. ROOTS \cite{laurencon2022roots}, the dataset that BLOOM \citep{scao2022bloom} was trained on, was styled after the Pile and would potentially be a good candidate, but it was not publicly available when we started training our models.
    \item As this framework is intended to be used as a baseline for future research, we feel it is important to stay close to currently accepted common practices. While the Pile is widely used for training English-language models, there is no equally widespread multilingual dataset. In particular, ROOTS has not been used to train models beyond BLOOM. 
    \item We do not have access to a multilingual evaluation framework that is anywhere near as comprehensive as \citet{gao2021eval}.
\end{enumerate}

We train 2 copies of the \textit{Pythia} suite using identical architectures. Each suite contains 8 models spanning 8 different sizes. We train one suite of 8 models on the Pile, and the other on a copy of the Pile after applying near-deduplication with MinHashLSH and a threshold of 0.87, following the advice that LLMs trained on deduplicated data are better and memorize less of their data
\citep{Lee2021DeduplicatingTD}. After deduplication, the deduplicated Pile is approximately 207B tokens in size, compared to the original Pile which contains 300B tokens.

\subsection{Architecture}

Our model architecture and hyperparameters largely follow \citet{brown2020language}, with a few notable deviations based on recent advances in best practices for large scale language modeling \citep{black2022gpt,chowdhery2022palm,zeng2022glm}:
\begin{enumerate}
    \item \citet{brown2020language} describes using sparse and dense attention layers in alternation, while we follow all subsequent work and use fully dense layers for our models.
    \item We use Flash Attention \citep{dao2022flashattention} during training for improved device throughput.
    \item We use rotary embeddings introduced by \citet{su2021roformer} and now in widespread use \citep{black2022gpt,chowdhery2022palm,zeng2022glm} as our positional embedding type of choice.
    \item We use the parallelized attention and feedforward technique and model initialization methods introduced by \citet{gpt-j} and adopted by \citep{black2022gpt,chowdhery2022palm}, because they improve training efficiency and do not harm performance.
    \item We use untied embedding / unembedding matrices, as prior work has suggested that this makes interpretability research easier \citep{belrose2023eliciting}.
\end{enumerate}

\subsection{Training}

We train our models using the open source library GPT-NeoX \citep{gpt-neox-library} developed by EleutherAI. We train using Adam and leverage the Zero Redundancy Optimizer (ZeRO) \citep{rajbhandari2020zero} to efficiently scale to multi-machine set-ups. We additionally leverage data parallelism \citep{goyal2017dataparallel} and tensor parallelism \citep{megatron-lm} as appropriate to optimize performance. We use Flash Attention \citep{dao2022flashattention} for improved hardware throughput.

The most notable divergence from standard training procedures is that we use a much larger batch size than what is standard for training small language models. It is widely held \citep{mccandlish2018empirical,zhang2019algorithmic,kaplan2020scaling,brown2020language,hoffmann2022training} that using larger batch sizes is desirable, but that smaller LLMs require smaller batch sizes to avoid convergence issues. Contrary to this literature, we find no convergence issues with using batch sizes $4\times$ to $8\times$ what is considered standard for models with less than 1 billion parameters. Consequently, we use a batch size of 1024 samples with a sequence length of 2048 (2,097,152 tokens) for all models, in order to maintain consistency across all \textit{Pythia} model training runs.

\begin{table}[h]
    \centering
    \begin{tabular}{rccc}\toprule
     Model Size & GPU Count & GPT-3 GPUs & Speed-Up\\\midrule
          70 M  & 32 & 4 & 8$\times$\\
         160 M  & 32 & 8 & 4$\times$\\
         410 M  & 32 & 8 & 4$\times$\\
         1.0 B  & 64 & 16 & 4$\times$\\\bottomrule
    \end{tabular}
    \caption{Models in the Pythia suite, number of GPUs used during training, and the number of GPUs we would have been able to use had we used the GPT-3 suite's batch sizes. Due to the ability of GPT-NeoX to scale linearly as the number of GPUs increases, this produces substantial wall-clock speed-ups for small models. All GPUs are A100s with 40 GiB VRAM.}
	\label{table:hardware}
\end{table}

A large batch size is essential to training models quickly: in a regime where one is not bottlenecked by access to GPUs or high quality interconnect, doubling the batch size halves the training time. A maximum batch size therefore directly implies a \textit{minimum} wall-clock training time and \textit{maximum} number of compute-saturated GPUs. By inflating batch sizes beyond previous standards, we achieve wall-clock speed-ups of factors as large as $10\times$ compared with standard batch sizes on our smaller models (\cref{table:hardware}). We also note that our models still perform on par with widely used models of the same size like GPT-Neo \citep{black2021gpt} or OPT \citep{zhang2022opt} (see \Cref{app:evals} for plots on common benchmarks).

We save model checkpoints at initialization and every 2,097,152,000 tokens (or 1,000 iterations), resulting in $144$ checkpoints evenly spaced throughout training. Additionally, we save log-spaced checkpoints early in training at iterations $\{1,2,4,8,16,32,64,128,256,512\}$. This gives a total of $154$ checkpoints per model, far more than any other suite of publicly available language models.

We train all models for 299,892,736,000 $\approx$ 300B tokens, token-matching our models to the original GPT-3 and OPT model suites. The standard (duplicated) Pile is 334B tokens using the GPT-NeoX tokenizer, so some data in the Pile may not be seen by the standard Pythia models. For this reason we urge anyone seeking to study the effect of training data on the Pythia models use our provided data loaders to ensure accurate counts. The deduplicated Pile only contains 207B tokens, so we run for $\approx$1.5 epochs on it. This allows users of the Pythia suite to study deduplication in greater detail by comparing models shortly before the epoch boundary to those slightly after the epoch boundary. We find that there is no evidence that the second epoch negatively impacts evaluation scores on a variety of benchmarks (for more information, see \Cref{sec:novel} and \Cref{app:evals}).

We refer to the models trained on the original Pile as ``Pythia-xxx'', where `xxx' is the model's total parameter count rounded to 2 significant figures, and their counterparts trained on the deduplicated Pile as ``Pythia-xxx-deduped''.

\subsection{Evaluation}

While the primary focus of this work is to promote scientific research on the behaviors of large language models, and state-of-the-art performance is not necessarily a core requirement, we find that Pythia and Pythia (Deduplicated) perform very similarly to OPT and BLOOM models on a variety of NLP benchmarks. These results are presented in \cref{app:evals}. We use the Language Model Evaluation Harness \citep{gao2021eval} to run evaluations on eight common language modeling benchmarks (\cref{app:evals}). 
We consistently find that Pythia and Pythia (Deduplicated) perform very similarly to OPT and BLOOM models.

\subsection{Novel Observations in Evaluation}\label{sec:novel}

We find three interesting phenomena that run counter to the prevailing narratives in the literature. Firstly, we find that deduplication of our training data has no clear benefit on language modeling performance. This is consistent with the results of \citet{black2022gpt}, but inconsistent with other papers. This may indicate that the upsampling of certain subsets of the Pile does not accord with conventional assumptions about duplicated data, or that the general tendency of deduplicated data to outperform non-deduplicated data is primarily a statement about the quality of the data used in other works. Secondly, we find that we achieve (equi-token and equi-parameter) performance on-par with OPT despite the use of parallel attention + MLP sublayers at all model scales. Both \citet{black2022gpt} and \citet{chowdhery2022palm} state that this architecture choice causes a performance regression at scales $<$ 6B parameters. Thirdly, we find a minimal and inconsistent ``curse of multilinguality'' \citep{goyal2020unsupervised,pfeiffer2022lifting} for BLOOM. While BLOOM certainly underperforms other models on LAMBADA, PIQA, and WSC, it does not appear to do so on WinoGrande, ARC-easy, ARC-challenge, SciQ, and LogiQA. We interpret this as a sign that some of the existing literature on the curse of multilinguality may need to be revisited using more diverse evaluation benchmarks. Plots supporting all of these claims can be found in \cref{app:evals}.

\subsection{Public Release and Reproducibility}


To ensure that our work is fully reproducible, we seek to only make use of codebases and dependencies that are freely and publicly available. As previously mentioned, we use the open source GPT-NeoX and DeepSpeed libraries for training. For evaluating our models we use the Language Model Evaluation Harness \citep{gao2021eval} and run all evaluations ourselves instead of copying claimed results from previous papers.

We release all of our models and checkpoints to the public under the Apache 2.0 license via the HuggingFace Hub \citep{wolf2019huggingface}\footnote{\url{https://huggingface.co/EleutherAI}}
We additionally release the code used for all evaluations and the raw benchmark scores generated on GitHub.\footnote{\url{https://github.com/EleutherAI/pythia}}

In addition to training our models on the public Pile dataset,
 we also provide a tool for downloading the pre-tokenized data files utilized by our dataloader in the GPT-NeoX library, as well as a script that can be used to reproduce the exact dataloader used by our models during training, so that the contents of each batch at each training step can be read out or saved to disk by researchers.

\section{Case Studies}\label{sec:case-studies}

We perform three case studies in language modeling research that would not have been possible to perform using any pre-existing model suites. These case studies were chosen to cover a variety of topical domains and address small but important questions in their respective fields. We especially seek to leverage the public training data order to derive novel insights about these models that have not been previously studied.

\subsection{How Does Data Bias Influence Learned Behaviors?} \label{bias_section}
Large language models are typically trained on minimally curated human-authored data. While it is widely known that models typically learn the  biases encoded in their training data, virtually nothing is known about the actual learning dynamics of how these biases develop throughout training. This is particularly concerning as one of the best established phenomena in the study of bias in deep learning models is \textit{bias amplification}---the fact that social biases in deep learning models tend to be more extreme than those found in their training data \citep{zhao2017men,hirota2022quantifying,hall2022systematic}. 
To mitigate the biases learned from data, previous works have used finetuning on balanced datasets to reduce the gender bias of language models with some success \citep{levy2021collecting,gira2022debiasing,kirtane2022efficient}, yet little is known about the role of specific corpus statistics in the emergence of bias during pretraining.

We seek to investigate a counterfactual claim---if we were to train our models on a corpus with different properties, how would these models' properties change downstream? To test the effects of corpus statistics on the biases learned by language models, we repeat segments of pretraining on specific models, with altered corpus statistics. In particular, for the size 70M, 410M, 1.4B, and 6.9B Pythia (deduplicated) models, we take a checkpoint and optimizer state 21B tokens (7\%) prior to the end of training, and resume training of the model such that it sees the exact same data until the end of training, but with morphologically masculine pronouns replaced by their feminine counterparts. We also repeat this intervention for 63B tokens (21\%) prior to the end of training on just the Pythia-1.4B-deduped model. We then measure model performance on the WinoBias~\citep{zhao2018gender} benchmark and the English subset of the multilingual CrowS-Pairs~\citep{neveol-etal-2022-french}\footnote{While previous works have found the original version of CrowS-Pairs~\citep{nangia-etal-2020-crows} benchmark of questionable validity \citep{blodgett-etal-2021-stereotyping}, \citet{neveol-etal-2022-french} have revised the English dataset to take care of the raised concerns.} to observe whether this altered pretraining data affects downstream gender bias.
Neither of these benchmarks were originally intended for autoregressive language models or text generation, so we describe our modifications to the evaluation setups in \Cref{app:biasevals}.

The controlled setup provided by Pythia---with precise access to the data samples seen during training---enables us to isolate the effect of pronoun frequency in pretraining. If instead we chose to compare two different training datasets, we would change a large number of potential explanatory factors that we cannot control for. In fact, it has been suggested that even the choice of hyperparameters, such as the data ordering, can have an effect on the resulting bias~\citep{d2020underspecification}. Therefore, without being able to resume pretraining on the exact same data in the exact same order, we could not be confident our experiment was indeed measuring only the effect of particular gendered terms' frequency.

\begin{figure}[!ht]
\centering
\includegraphics[width=\linewidth, clip]{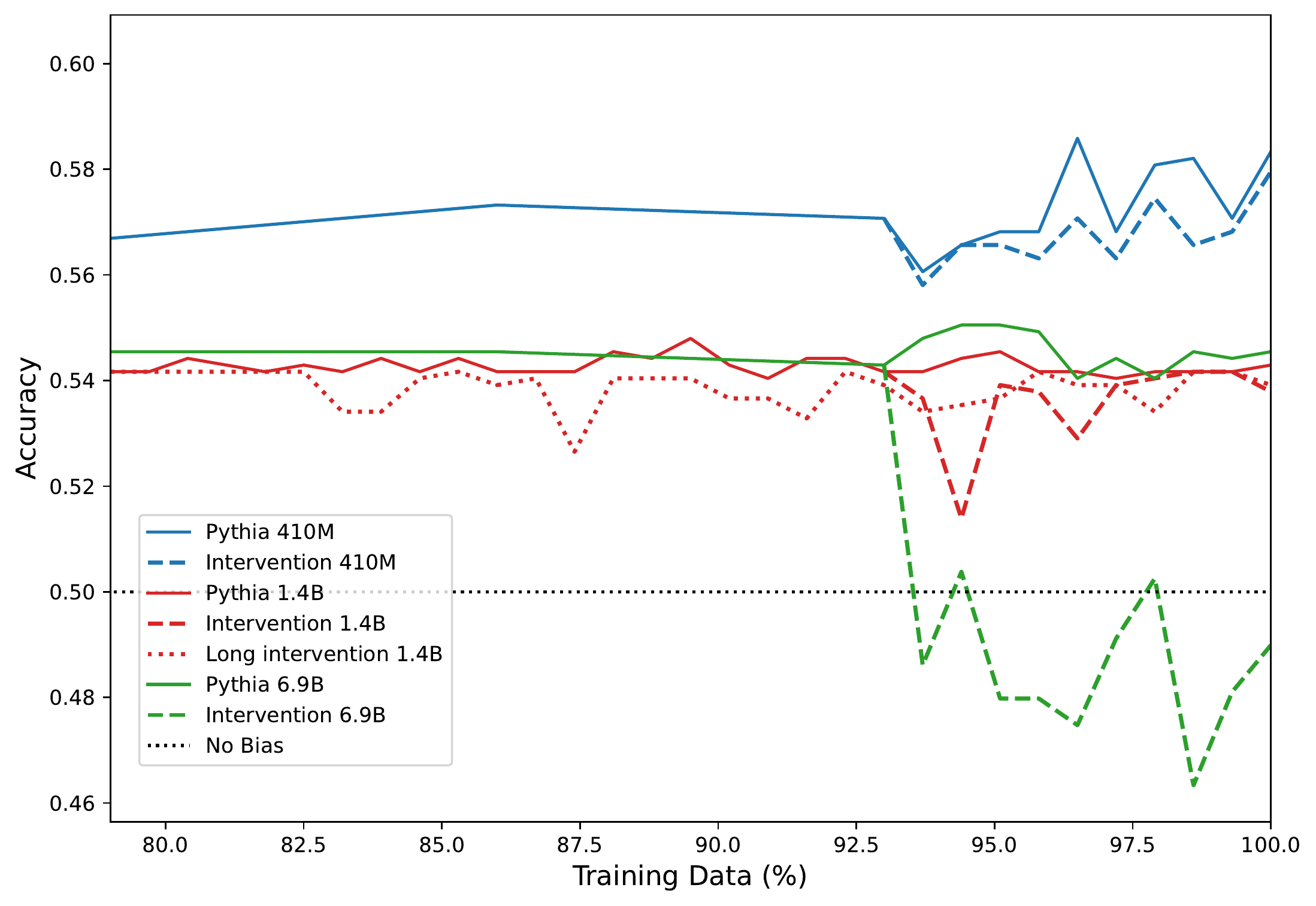}
\caption{The WinoBias gender bias results, shown as the proportion of the time that the model placed a higher log probability on the more stereotyped pronoun as an answer to a multiple choice gender--occupation co-reference question.}
\label{figure:winobias_mc}
\end{figure}

For our WinoBias implementation (see \Cref{app:biasevals}), we see a clear effect of the intervention in \cref{figure:winobias_mc}: a decrease in stereotypical accuracy for each intervention and across model scale. On the largest model scale tested, 6.9B, applying the intervention also successfully changes the model throughout training on the intervention from a pro-stereotypical bias to an anti-stereotypical one. We hypothesize that these results indicate that larger capacity models show less pro-stereotypical bias due to their ability to learn more complex relationships between occupation and pronouns, and that the intervention effect size increases across scale for similar reasons.


\begin{figure}[!ht]
\centering
\includegraphics[width=\linewidth, trim = {1cm 1cm 3cm 2cm}, clip]{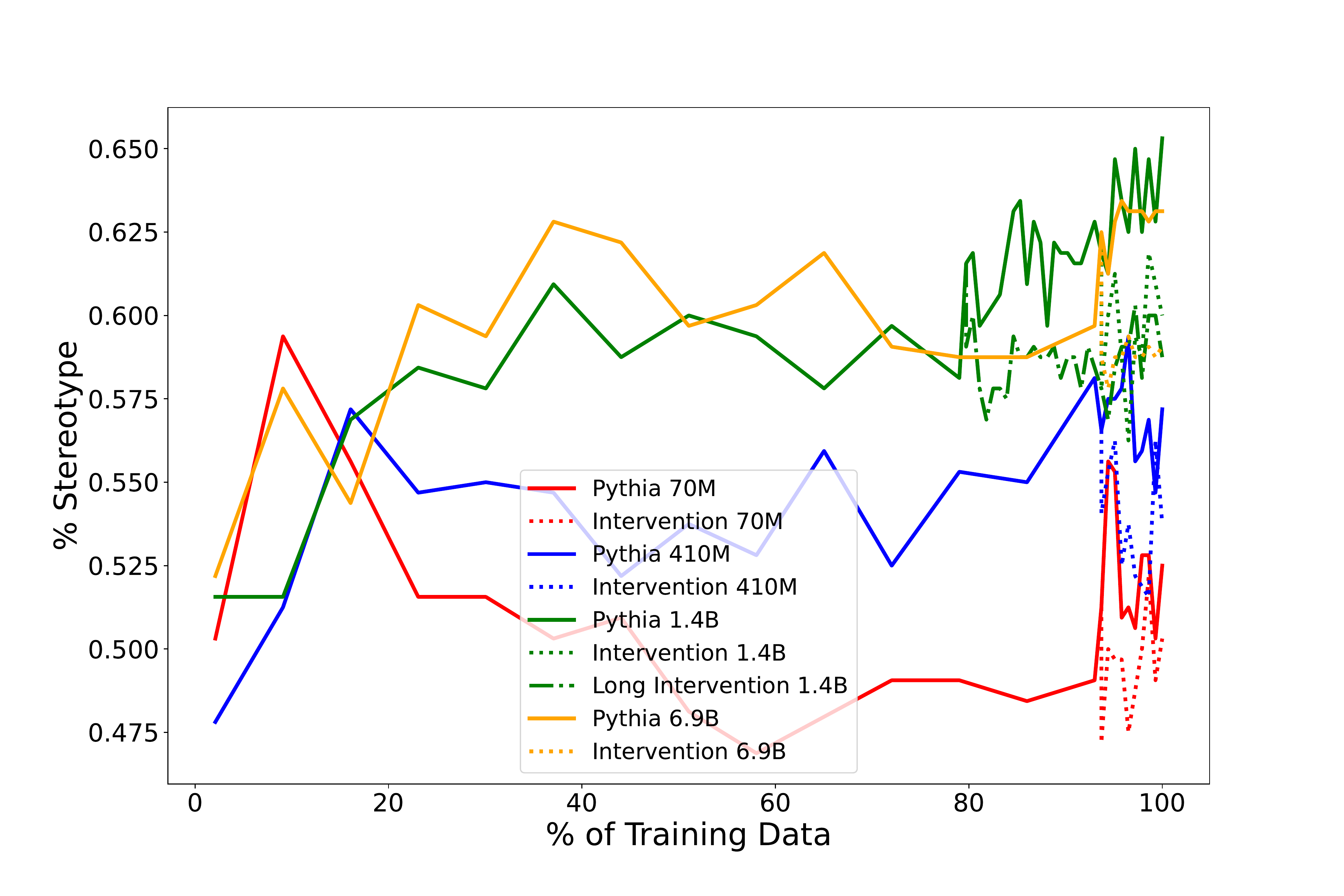}
\caption{The CrowS-Pairs gender bias, shown as the percentage of times that the perplexity of the stereotyping sentence is lower than its less stereotyped counterpart (\% Stereotype) for the Pythia models of different sizes at the end of training. We also show the effect of the gender swapping intervention on the measured bias for the partially retrained models.}
\label{figure:crowspairs}
\end{figure}

\cref{figure:crowspairs} shows the progression of the CrowS-Pairs gender bias metric and the effect of the interventions. We can clearly see a reduction in the bias as a result of swapping the gendered pronouns in the last 7\% or 21\% of the training for all model sizes, but most prominently for the larger ones, although these are also more biased to begin with. We hypothesize that because larger models are better at modeling correlations and distributions within their corpora, their increased capacity causes features of bias to be more strongly or robustly learned.
We also see that the interventions only lead to a marginal decrease in the model perplexity on LAMBADA~\citep{paperno2016lambada} (\cref{app:biasevals}), which demonstrates the effectiveness of the bias mitigation without hurting language modeling performance downstream to a large degree.
Whether the noisiness of the progression reflects actual changes in the language model's bias or poor reliability of CrowS-Pairs is an open question we leave for future work.


We propose that performing such modifications to portions of language model training data, retraining, and comparing to the baseline model (``interventions'') should be studied further for applications including but not limited to investigating bias amplification and devising new mitigation strategies.
For example, while not explored in this case study, we think that the finegrained information that Pythia provides on the data seen during training could benefit the promising literature on influence functions to estimate the role of specific training samples on the encoded bias~\citep{brunet2019understanding,silva2022cross}.
While this was beyond the scope of this case study, we believe that the extensive availability of checkpoints, consistent training order, and retrainability could be useful in assessing the \textit{test-retest reliability} of existing bias measures \citep{van2022undesirable}.


\subsection{Does Training Order Influence Memorization?}

Although memorization in neural language models is widely studied, many basic questions about the dynamics of memorization remain unanswered. Prior work on the dynamics of memorization is generally limited to a few models in isolation \citep{jagielski2022measuring,elazar2022measuring} or papers which train (but do not release) custom models for their studies \citep{tirumala2022memorization,hernandez2022scaling}. \citet{carlini2022quantifying} studies the impact of scaling on memorization and repeatedly remark on the lack of suitable model suites for their study. They ultimately focus on the GPT-Neo model suite \citep{black2021gpt,gpt-j,black2022gpt}, despite the fact that these models were trained on slightly different datasets, in different orders, and with inconsistent checkpointing. 

In this experiment we test whether training order influences memorization. This is an explicitly theoretically-driven experiment: several authors realized that their mental model of transformers was that they work iteratively---by adding new information to a latent space and then processing the space as a whole to obtain a better representation. This mental model predicts that data encountered later in training will be memorized more, as the model has had less time to incorporate it more fully into its representation space. If true, this would potentially be highly useful for mitigating the memorization of sequences for which verbatim memorization would be undesirable, by intentionally modifying a model's training data order prior to training.

To test our hypothesis, we measure the memorization of an initial segment of each sequence in the training corpus. While there are several reasonable definitions of memorization, we use the one from \citet{carlini2021extracting} as it has received considerable attention in the literature \citep{yoon2021model,huang2022large,ginart2022submix,ippolito2022preventing,biderman2023emergent}. In their context, a string is $(k,\ell)$-memorized if prompting the model with a string of length $k$ from the training data induces the model to generate the next $\ell$ tokens from the training data correctly. We choose $k=\ell=32$ largely arbitrarily, and note that doing all reasonable pairs of $(k,\ell)$ would have a computational cost comparable to retraining all of our models from scratch. To avoid potential covariate effects, we only use the first 64 tokens from each context seen during training.

\begin{figure}[!htb]
    \centering
    \includegraphics[width=\columnwidth]{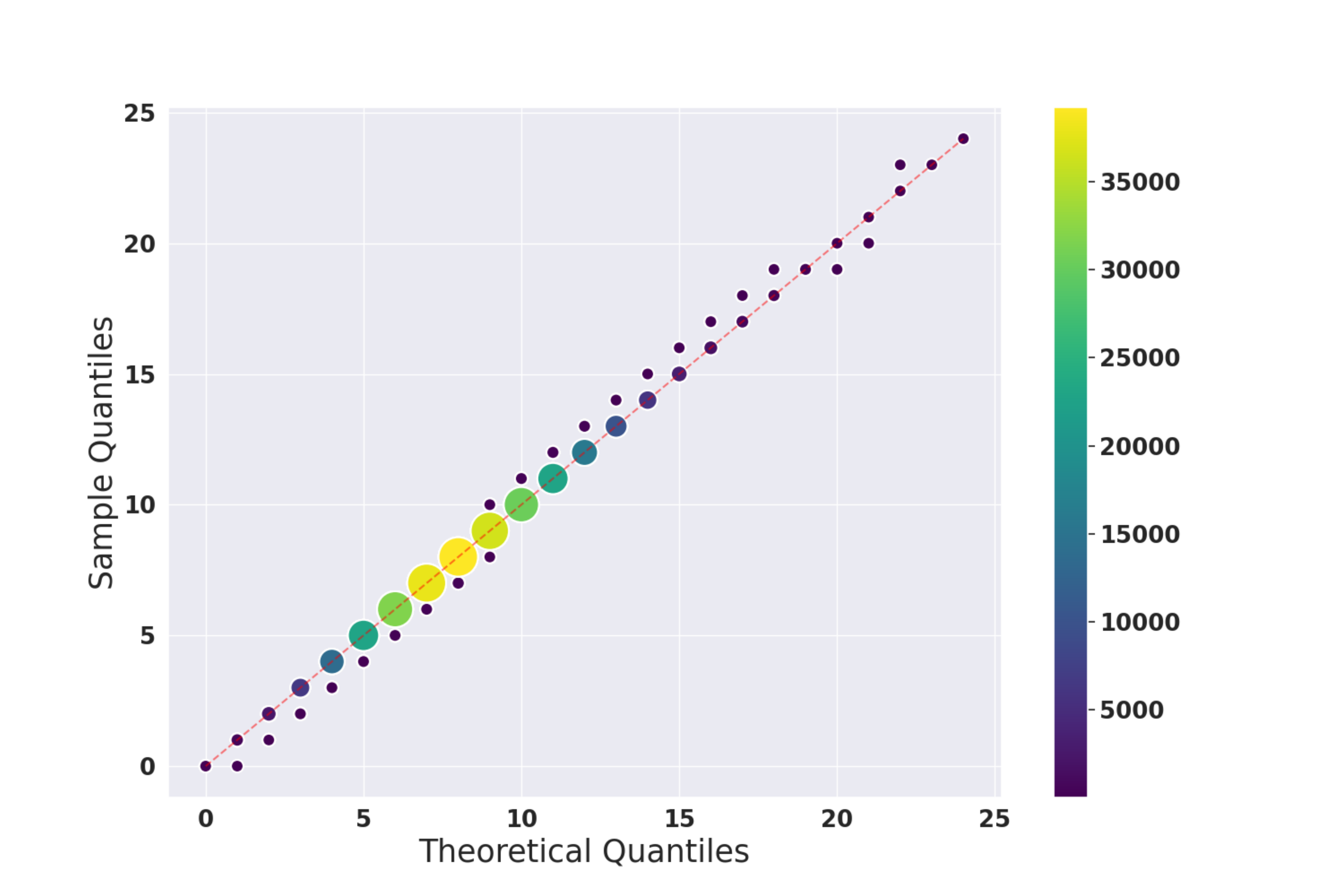}
    \includegraphics[width=\columnwidth]{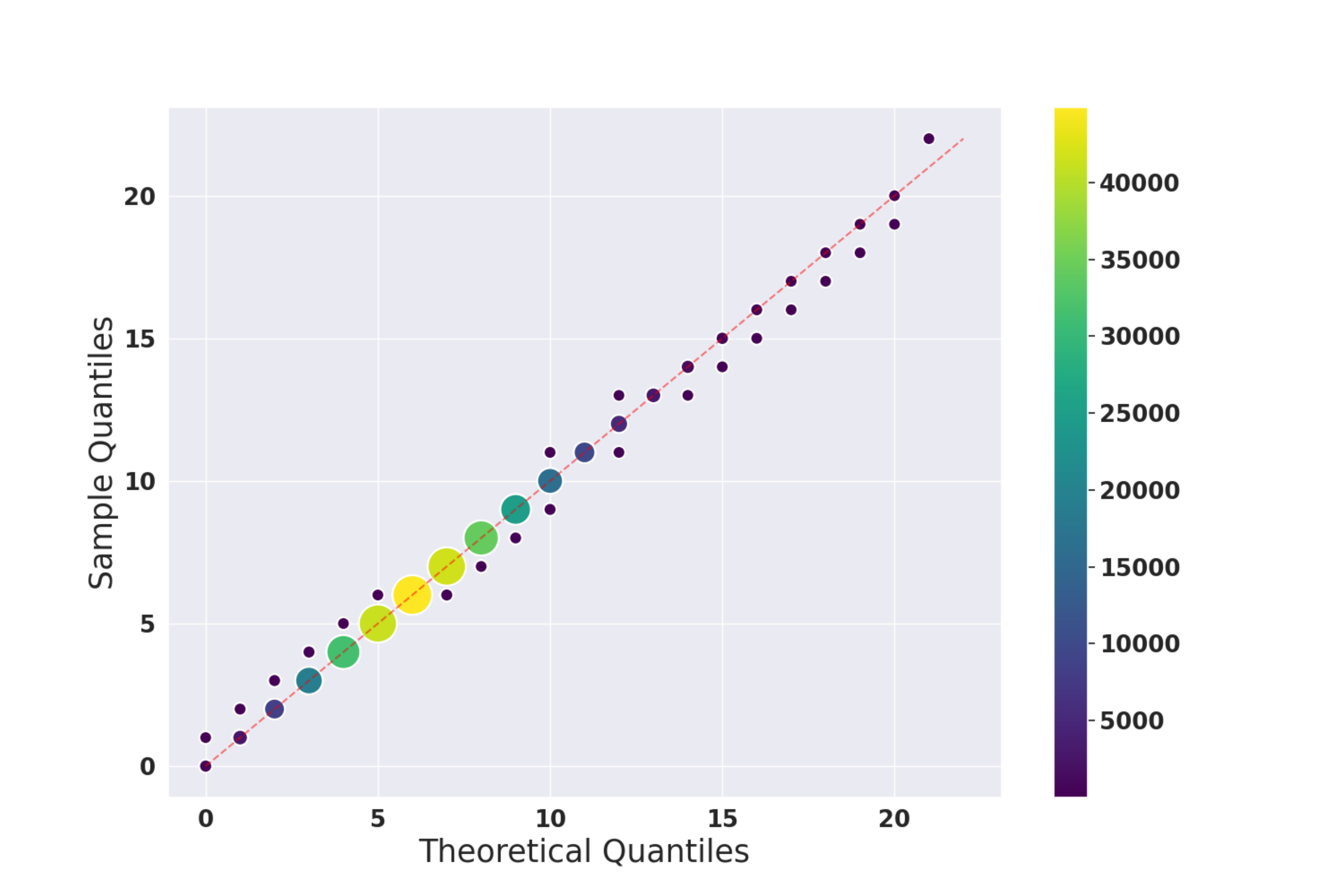}
    \caption{Quantile-Quantile plot of rate of occurrence of memorized sequences in 12B model compared to a Poisson Point Process, with (top) and without (bottom) deduplication. Color and dot size indicates number of points. We assume each mini-batch to be a time-slice in a Poisson process where we count the events (number of memorized sequences) within a time-slice.}
    \label{fig:qq_plots}
\end{figure}

Surprisingly, we find that a Poisson model fits the data extremely well (\cref{fig:qq_plots}), indicating that training order has little impact on memorization. This model implies that memorized sequences are not spaced more densely toward the beginning or end of training, and that between each checkpoint roughly the same number of memorized sequences can be found.

The Poisson process here describes an event of the occurrence of a memorized sequence within a batch of training data. As the evaluation was performed on the first 64 tokens of every sequence within the training corpus, in the same order of training, we can consider each batch to represent a hypothetical time interval, where a unit of time corresponds to a sequence of the training corpus, with sample distribution defined as the number of memorized sequences in a batch of training data, and the theoretical distribution as the best fit Poisson distribution from samples. We use a batch size of 512 sequences for these plots, but we observe similar results for various batch sizes.


The count (color bar to the right in \cref{fig:qq_plots}) indicates the density of plotted points (also indicated by size) on the Q-Q plot. Q-Q plots serve the purpose of being a goodness of fit test for asserting the fact that the rate of occurrence of memorized sequences in training data is uniform.


This finding is important for practitioners seeking to control which sequences are memorized by a model. It implies that one cannot simply place sequences that are undesirable to memorize at the beginning or end of training and successfully reduce the chance of memorization. However, we propose that a practitioner especially worried about the memorization of certain sequences place those sequences at the beginning of training, thus increasing the odds that the practitioner may observe prior to the completion of the training run that undesirable memorization behavior occurs in the partially-trained model.

\subsection{Do Pretraining Term Frequencies Influence Task Performance Throughout Training?}


\begin{figure*}[!htb]
\centering
\includegraphics[width=\textwidth]{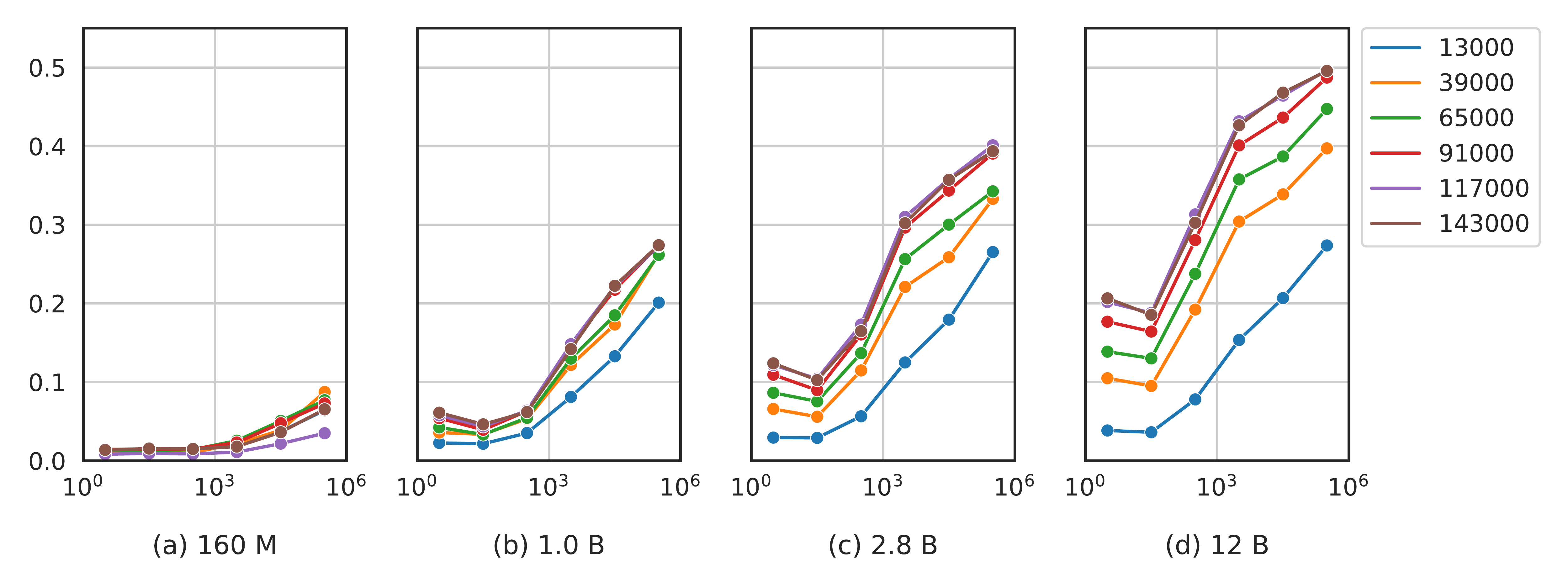}
\caption{Accuracy on Trivia QA plotted againts the number of relevant entity counts found in a QA-pair. Each subfigure shows the impact of performance across various model sizes over multiple intermediate checkpoints. (With train step counts denoted by color on the right) Each point represents the average accuracy ($y$-axis) of binned counts ($x$-axis).}
\label{figure:trivia_qa}
\end{figure*}

Recent work has explored the effect of statistics of language model corpora on numerous downstream tasks. Findings presented in \citet{shin2022effect} demonstrate how the pretraining corpus can impact few-shot performance, while \citet{razeghi2022impact} investigates how models are able to perform numerical reasoning from in a few-shot setting. By charting the performance of a arithmetic task given an input operand and the frequency at which it is found in the pretraining corpus, they concluded that accuracy tends to be higher for terms that are found more frequently compared to terms that are less frequent. Other works also suggest that the pretraining corpus has a significant impact on few-shot behavior \citep{elazar2022measuring,kandpal2022large}. These works observe a correlational and causal relationship between the ability to answer factual questions and the frequency of salient entities found in the pretraining corpus. While the aforementioned works experiment with various model sizes, it is not yet studied when during training and at what model sizes this effect occurs. We further investigate this phenomenon across model checkpoints and model sizes by adapting arithmetic tasks of multiplication and addition \citep{razeghi2022impact} and a QA task \citep{kandpal2022large} using natural language prompts evaluated over a set of $k$-shot settings. We calculate the relevant term frequencies for all model checkpoints based on the pretraining data seen by each checkpoint, which means counting through each subset of the pretraining corpus sampled and seen by the model up to each chosen train step. Model evaluation was performed on the \textit{Pythia} (Deduplicated) suite using the LM Evaluation Harness \cite{gao2021eval}.

Following \citet{razeghi2022impact}, the formulation of the arithmetic task consists of input operands $x_1 \in [0,99]$ and $x_2 \in [1,50]$ and an output $y$. The input operands are converted into a prompt with the prompt template \textit{``Q:What is $x_1$ \# $x_2$? A:''} with \# being \textit{``plus''} for addition and \textit{``times''} for multiplication. We measure the accuracy of a prompt instance by checking the model's prediction against $y$. To measure the term frequency and task performance correlation, the average accuracy of all prompts with the same $x_1$ over all values of $x_2$ is mapped to the number of times $x_1$ is found in the sampled pretraining data that each evaluated model checkpoint sees. In few-shot settings, we sample examples with digits that differ from the $x_1$ values being measured. 

As a QA task, we use TriviaQA \citep{joshi2017triviaqa} with a simple template of \textit{``Q: $x_1$ \textbackslash n A: $y$''} with $x_1$ being the question and $y$ answer, where $y$ is included for a few-shot sample or left blank for the sample being evaluated. The model prediction is evaluated with exact match over the set of possible answers. The term frequencies of a single question-answer pair (``QA pair'') are calculated based on the number of times all salient entities of that QA pair appear in a sampled pretraining data sequence seen by a given checkpoint. We follow the original experiment using 4 shots and evaluate both the training and the validation split of the dataset. Performance is averaged over a group of log-spaced bins. 

We observe that for both arithmetic and QA experiments, model sizes affect the correlation between average performance and the term frequencies, indicating that this correlation is an emergent property in larger models. Smaller models rarely produce accurate results on the task despite being given up to 16 few-shot examples, as shown in \cref{figure:addition}, where models at sizes below 1 billion are unable to perform well even in later stages of training, suggesting that these models are not successful at learning these tasks regardless of frequency of pertinent information in their training data. Similar patterns can be seen in \cref{figure:trivia_qa} where performance increase as training progresses mainly happens for larger models only. For the multiplication task, we also calculate the performance discrepancy between the top 10\% most frequent input operands and the bottom 10\% least frequent input operands also following \citet{razeghi2022impact} (see \cref{table:performance-gap}). We find that this performance gap widens over the course of training.

\textit{Pythia} allows the observation of the dynamics of which term frequencies affect performance in greater clarity than previous works. With confounding factors such as difference in model architecture, pretraining datasets, and training hyperparameters removed, we can better understand when effects that term frequencies have over a model's task performance occur. In practice, observing the phenomenon with respect to model size and intermediate checkpoints allows for better choices in future training runs. For example, if one cares about a model knowing the answer to some given question, one can calculate how many times that information occurs in the training data to predict whether it is likely or less likely a model of X size will be capable of retaining and recalling this information from its training data.


\section{Conclusion}\label{sec:conclusion}

We release \textit{Pythia}, a suite of language models trained with consistent data ordering and model architecture across multiple orders of magnitude of scale. We demonstrate how Pythia can be used to empower experiments at unprecedented levels of detail for a public model suite by presenting novel analyses and results on gender debiasing, memorization, and term frequency effects. We hope that these analyses will inspire further follow-up work showing how pretraining data drives the acquisition and emergence of capabilities across more complex tasks and that these models and their dataset tooling will be broadly useful for a variety of practitioners, and recommend using the suite as a framework for novel experimental setups on LLMs.

\section*{Acknowledgments}

We are grateful to Stability AI for providing the compute required to train these models, and to CoreWeave for providing compute for some of the evaluations.
OW's contributions are financed by the Dutch Research Council (NWO) as part of project 406.DI.19.059. HB's contributions were supported by the UKRI Centre for Doctoral Training in Application of Artificial Intelligence to the study of Environmental Risks (reference EP/S022961/1).

We thank Nora Belrose, Tim Dettmers, Percy Liang, Yasaman Razeghi, Mengzhou Xia, and various members of the EleutherAI Discord Server for their feedback.

We also thank the developers of the GPT-NeoX, Megatron-DeepSpeed, and NeMo Megatron libraries for their assistance and support, and Vincent Hellendoorn for contributing the implementation of Flash Attention, enabling us to save substantial time training the models.

\bibliography{main}
\bibliographystyle{icml2023}

\newpage
\appendix 
\onecolumn

\section{Author Contributions}

All authors other than the first two are listed in alphabetical order.

\paragraph{Stella Biderman} Conceived, organized, and lead the project. Designed the experiments for the memorization and pretraining frequencies case studies. Lead the writing of the paper.

\paragraph{Hailey Schoelkopf} Trained the models, wrote the paper, uploaded and converted all model checkpoints for hosting, and planned the gender bias case study. 

\paragraph{Quentin Anthony} Optimized the model implementation, advised the choice of hyper-parameters, and wrote the paper.

\paragraph{Herbie Bradley} Carried out the WinoBias analysis and wrote portions of the gender bias case study.

\paragraph{Kyle O'Brien} Conducted zero- and five-shot evaluations of several of the models on NLP benchmarks.

\paragraph{Eric Hallahan} Evaluated the models on standard NLP benchmarks and authored most plots in the paper.

\paragraph{Mohammad Aflah Khan} Helped in implementing the CrowS-Pairs evaluation and performed analysis on the results.

\paragraph{Shivanshu Purohit} Optimized the model implementation, advised the choice of hyperparameters.

\paragraph{USVSN Sai Prashanth} Conducted the memorization case study, evaluated the models on standard NLP benchmarks and wrote the paper.

\paragraph{Edward Raff} Advised on the project and wrote the paper.

\paragraph{Aviya Skowron} Wrote documentation for the model suite and analysis, including the model card. Edited the paper.

\paragraph{Lintang Sutawika} Conducted the experiments and wrote the section for the pretraining frequencies case study.

\paragraph{Oskar van der Wal} Helped with the CrowS-Pairs evaluation and writing up the gender bias case study.

\section{Corrections and Updates}\label{app:corrections}

Following the value of ``doing science in the open'' \citep{phang2022eleutherai}, we released a variety of artifacts over the course of training our models for the public to use. However, after this initial release of preliminary versions of the \textit{Pythia} suite (``\textit{Pythia} v0''), we decided that in order to make \textit{Pythia} as controlled as possible, it was necessary to update the model suite with slightly better-controlled hyperparameter selection.

The updated version of the \textit{Pythia} suite (``v1'') features several small changes to hyperparameters in a redone version, detailed below:

\begin{itemize}
    \item All model sizes are now trained with uniform batch size of 2M tokens. Previously, the models of size 160M, 410M, and 1.4B parameters were trained with batch sizes of 4M tokens, but in the course of training the initial suite we discovered that it was feasible to train all models with uniform batch size, though based on prior literature we had not been certain of this fact before performing our own experiments on batch size.
    \item We configured additional model checkpoint saving in order to obtain checkpoints at initialization (step 0) and steps~$\{1,2,4,8,16,32,64,128,256,512\}$ in addition to every 1000 training steps. This enables practitioners to use our new suite to study training dynamics and emergent behaviors early in training, as well as access the random weight initializations easily.
    \item Before retraining the suite, we received a contribution to our codebase integrating Flash Attention \citep{dao2022flashattention}. Utilizing the Flash Attention fused attention kernel greatly increased per-device throughput for the second set of training runs.
    \item We remedied a minor inconsistency that existed in the original suite: all models of size 2.8B parameters or smaller had a learning rate (LR) schedule which decayed to a minimum LR of 10\% the starting LR rate, but the 6.9B and 12B models all used an LR schedule which decayed to a minimum LR of 0. In the redone training runs, we rectified this inconsistency: all models now were trained with LR decaying to a minimum of $0.1 \times$ their maximum LR.
\end{itemize}

We did not expect these changes to significantly impact any experimental findings in the paper, and we reran all analyses and evaluations on the new models to confirm this was indeed the case. All experiments in the paper report results from this updated version of the suite. We chose to rerun the training runs in order to make the \textit{Pythia} suite maximally useful to practitioners, and report this change for full transparency.

\begin{wrapfigure}{r}{0.5\textwidth}
    \centering
    \begin{tabular}{cc}\toprule
  Model Name & Previous Model Name\\\midrule
          70 M  & 19 M\\
         160 M  & 125 M \\
         410 M  & 350 M \\
         1.0 B  & 800 M \\
         1.4 B  & 1.3 B \\
         2.8 B  & 2.7 B \\
         6.9 B  & 6.7 B \\
         12 B  & 13 B \\
         \bottomrule
    \end{tabular}
    \caption{Model Names used for the \textit{Pythia} suite, before and after updating nomenclature to include the untied embedding / unembedding layers we use.}
	\label{table:name_change}
\end{wrapfigure}

We overwrote the previously public preliminary version of the suite (which now remains available at \url{https://huggingface.co/models?other=pythia_v0} to enable replicability of experiments using v0 of the suite) on March 31, 2023. Going forward, we will use semantic versioning for additional fixes as needed. Current best practices and details on further fixes can be found at  \url{https://www.github.com/EleutherAI/pythia}.

Additionally, on January 20, 2023, we chose to rename the \textit{Pythia} model suite to better reflect including both embedding layer and unembedding layer parameters in our total parameter counts, following the naming conventions from the GPT-2, BLOOM, and OPT suites, among others. We chose to do so to minimize documentation debt accrued in the field across model releases, and recommend future work explicitly use parameter counts derived from including embedding layers to obtain estimates more closely matching on-device memory required for running a given model. 

\section{Additional Plots for Case Studies}
\subsection{Gender Bias Interventions}\label{app:biasevals}


We also describe our modifications to the evaluation setups in the gender bias case study (see \Cref{bias_section}), as neither of the benchmarks were originally intended for autoregressive language models or text generation.

\paragraph{WinoBias} is a coreference resolution benchmark testing how a model links gendered pronouns to stereotypical occupations for each gender~\citep{zhao2018gender}. WinoBias contains both pro and anti-stereotypical versions of these tasks (the latter created by swapping pronouns), but we formulate the benchmark by taking only the pro-stereotypical subset and prompting the language model in multiple choice fashion with both pronouns, then obtaining log probabilities. To use this benchmark with our autoregressive language models, we use PromptSource~\citep{bach2022promptsource} to prompt our models with templates:
Given a sentence containing two occupations and a pronoun, the model is asked which of two pronouns an occupation refers to. We then take the pronoun with the highest \textit{log probability} and calculate a `stereotype accuracy' metric in which 1 represents perfectly predicting stereotypes and 0.5 represents random accuracy, or no bias.\footnote{For example, to query the model for an occupation linked with the pronoun `her', we might start with a sentence such as ``The mover greeted the librarian and asked the librarian where the books were.'', then append ``In this sentence, what can `the librarian' be replaced by: `him' or `her'? '' before prompting the model with the concatenation. The target completion for the model is then `her'.}
This formulation is different from the original WinoBias setup~\citep{zhao2018gender}, which measured the gender bias of older coreference approaches such as rule-based systems that do not require prompting.

\paragraph{CrowS-Pairs} is a stereotype benchmark that presents a model with two versions of a sentence: a stereotyped version and a version which is less stereotyping~\citep{neveol-etal-2022-french}. While the original task was designed for masked language models \citep{nangia-etal-2020-crows}, we measure the percentage of sentences for which the language model assigns a lower perplexity for the stereotyping sentence over the less stereotyping sentence. We evaluate our models only on the English subset for gender bias, since our models are monolingual and we intervene on gendered pronouns.

\begin{figure}[!ht]
\centering
\includegraphics[width=0.45\linewidth]{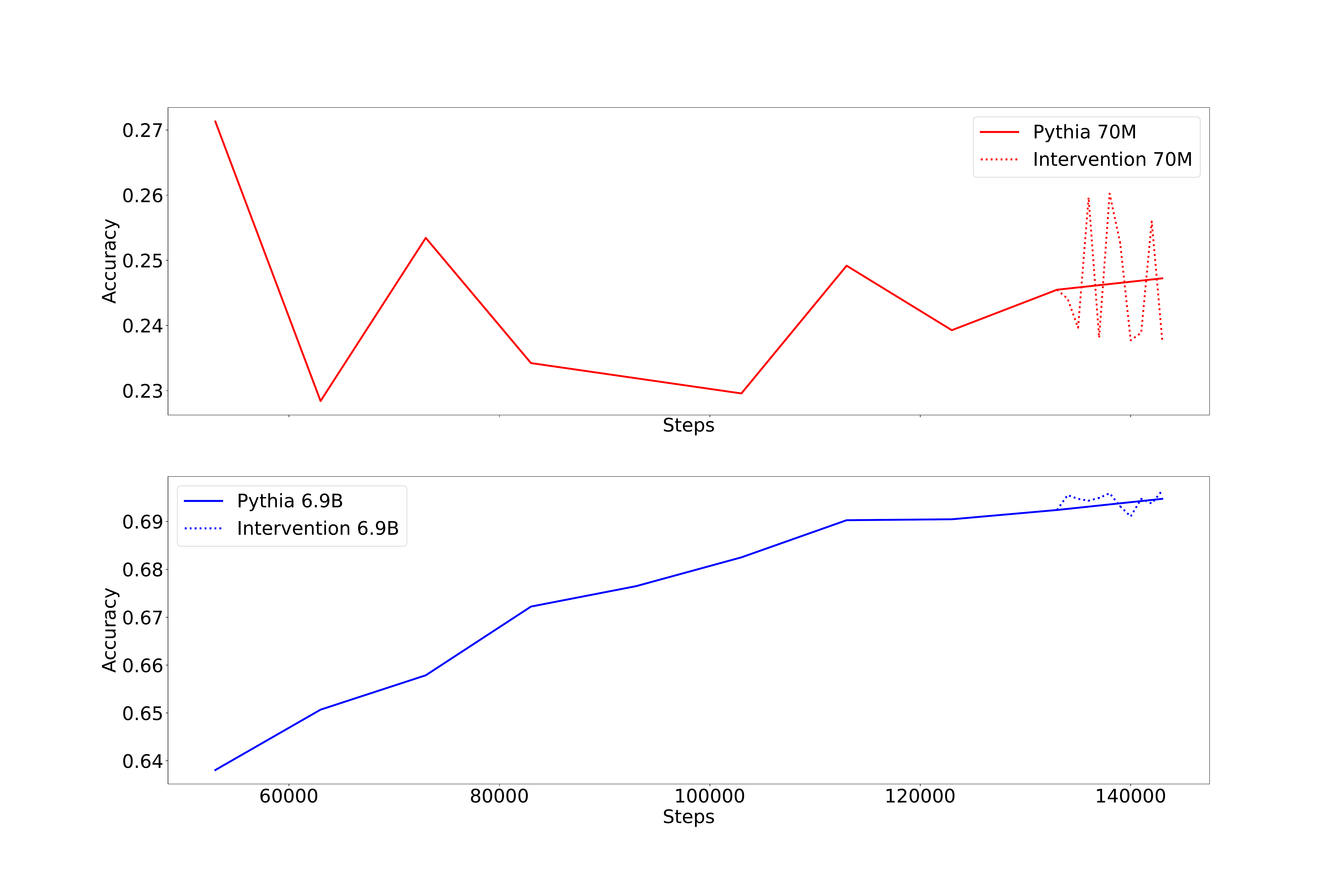}
\includegraphics[width=0.45\linewidth]{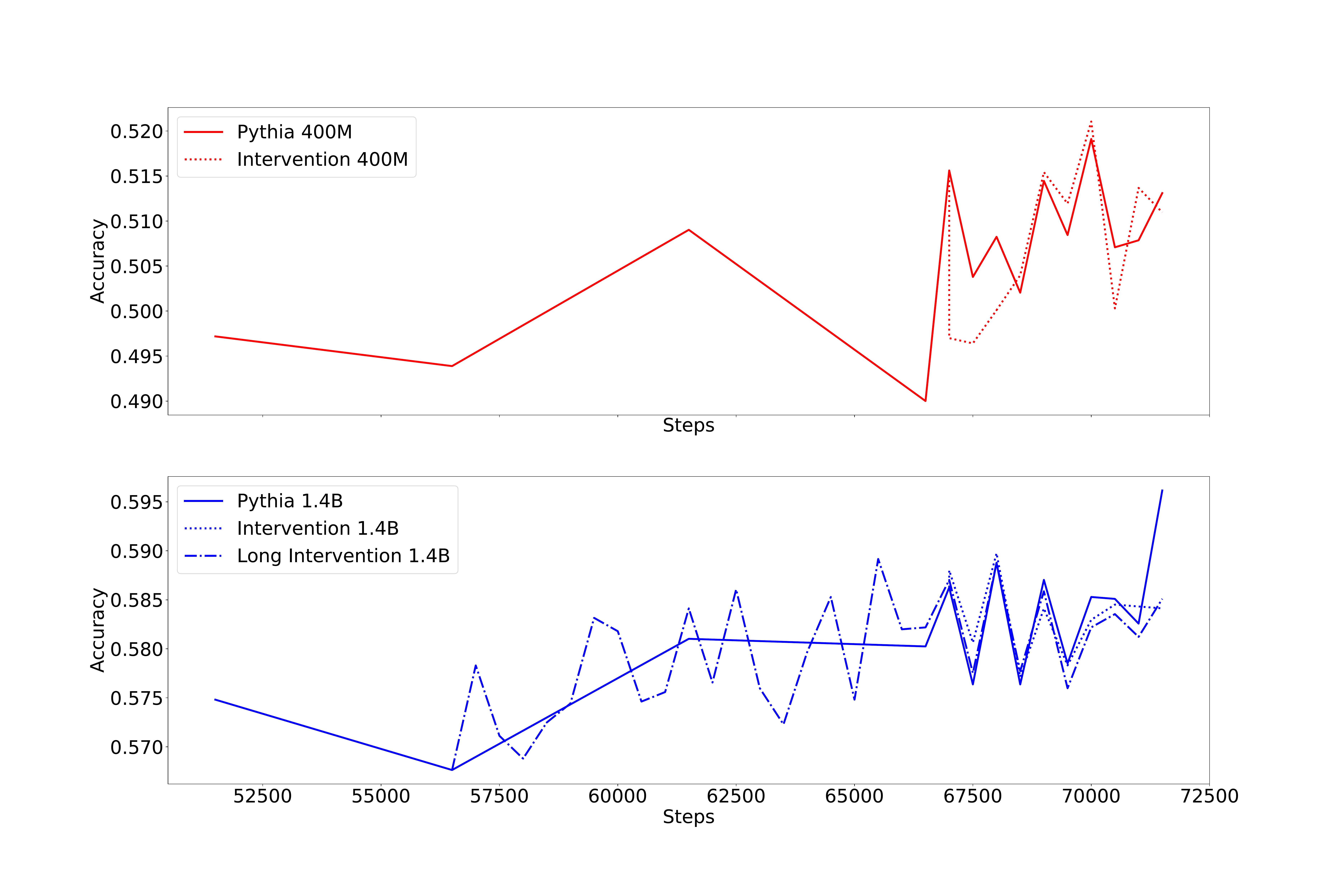}
\caption{Zero-shot evaluations of Pythia models over training, as well as their intervened counterparts, on the LAMBADA dataset.}
\label{figure:lambada_openai}
\end{figure}

\Cref{figure:lambada_openai} demonstrates the performance of different models in the \textit{Pythia} suite on the LAMBADA Dataset \citep{paperno2016lambada}. The plots also show how intervening by swapping gendered pronouns does not lead to major dips in accuracy. Hence the interventions are successful in reducing bias while preserving the text understanding capabilities of the model. 

\subsection{Pretraining Term Frequency}\label{app:freqevals}

\begin{table}[ht!]
\centering
\setlength{\tabcolsep}{2pt}
\begin{tabular}{l ccc ccc ccc ccc}
\toprule
\multirow{2}{*}{checkpoint} & \multicolumn{3}{c}{160 M} & \multicolumn{3}{c}{1.0 B} & \multicolumn{3}{c}{2.8 B} & \multicolumn{3}{c}{12 B} \\
\cmidrule(lr){2-4} \cmidrule(lr){5-7} \cmidrule(lr){8-10} \cmidrule(lr){11-13}
 & $\Delta_{k=0}$ & $\Delta_{k=4}$ & $\Delta_{k=16}$ & $\Delta_{k=0}$ & $\Delta_{k=4}$ & $\Delta_{k=16}$ & $\Delta_{k=0}$ & $\Delta_{k=4}$ & $\Delta_{k=16}$ & $\Delta_{k=0}$ & $\Delta_{k=4}$ & $\Delta_{k=16}$ \\
\midrule
13000  &  10.2 &   2.8 &  0.6 &  13.2 &   7.8 &   6.4 &   8.8 &  12.6 &  14.0 &   5.4 &  13.2 &  11.6 \\
39000  &   7.4 &   7.0 &  5.4 &  12.0 &  11.8 &  16.0 &   9.0 &  33.6 &  30.6 &  16.2 &  29.0 &  37.8 \\
65000  &   9.0 &   4.0 &  2.8 &  13.0 &  12.8 &  11.0 &  10.8 &  34.4 &  24.8 &  20.2 &  47.0 &  49.2 \\
91000  &  13.8 &  11.2 &  3.2 &  14.2 &  11.0 &  12.8 &   5.2 &  46.4 &  47.0 &  26.0 &  58.0 &  54.2 \\
117000 &   5.8 &   4.0 &  2.0 &  16.6 &  11.0 &  10.4 &   6.8 &  66.6 &  64.4 &  36.2 &  72.4 &  63.4 \\
143000 &  12.2 &   8.6 &  3.0 &  15.2 &  12.8 &  12.2 &   4.0 &  66.0 &  66.6 &  42.2 &  75.6 &  62.4 \\
\bottomrule
\end{tabular}
\caption{Performance gap on the arithmetic multiplication task for various model sizes with varying number of shots across checkpoints.}
\label{table:performance-gap}
\end{table}


\begin{figure*}[htb]
\centering
\includegraphics[width=\textwidth]{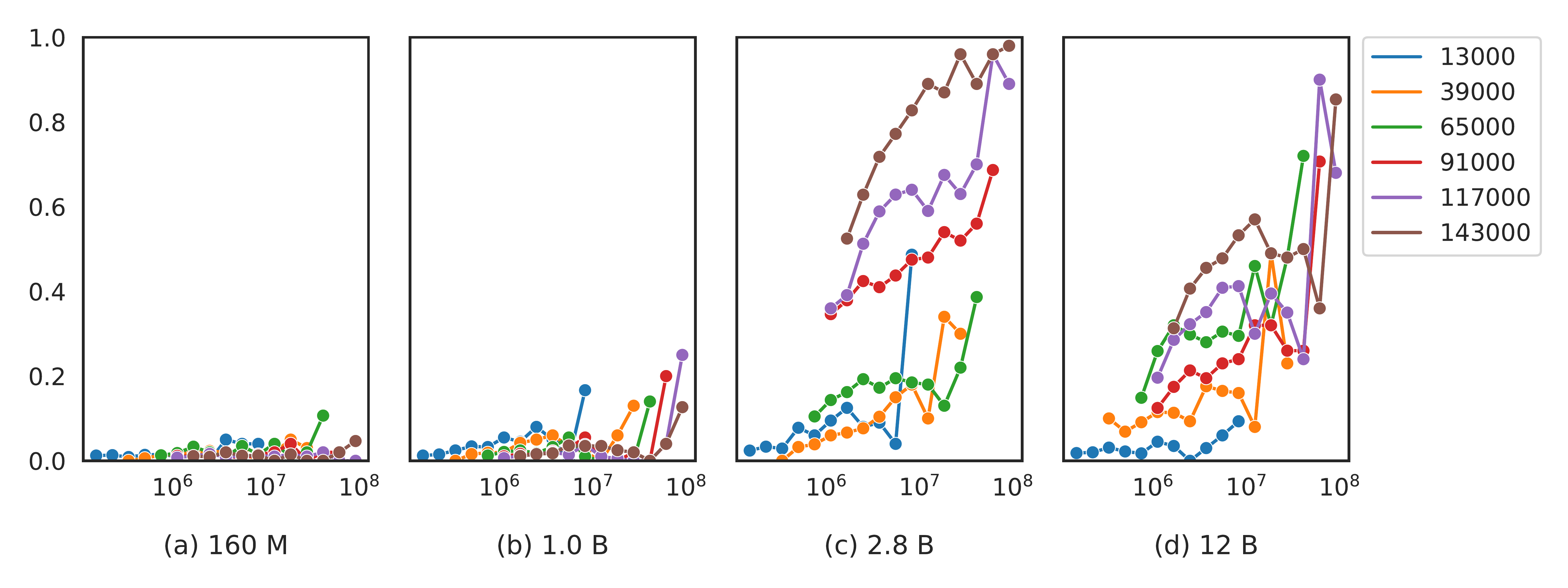}
\caption{Accuracy of the arithmetic addition task with 16 shots, across various model sizes (divided by subfigure). For each model, multiple intermediate checkpoints (differentiated by color and their step number) are plotted. Each point represents the average accuracy ($y$-axis) of binned term frequency ($x$-axis).}
\label{figure:addition}
\end{figure*}

\clearpage

\section{Training Hardware and GPU hours}\label{app:emissions}

We additionally report the number of accelerators used to train each \textit{Pythia} model size, alongside counts of total GPU-hours required for training our models at the throughputs that we achieve.

\begin{table}[h]
    \centering
    \begin{tabular}{rccc}\toprule
  Model Size & GPU Count & Total GPU hours required\\\midrule
          70 M  & 32 & 510 \\
         160 M  & 32 & 1,030\\
         410 M  & 32 & 2,540\\
         1.0 B  & 64 & 4,830 \\
         1.4 B  & 64 & 7,120 \\
         2.8 B  & 64 & 14,240\\
         6.9 B  & 128 & 33,500 \\
         12 B   & 256 & 72,300 \\\midrule
         Total   &  & 136,070 \\\bottomrule
    \end{tabular}
    \caption{Model sizes in the \textit{Pythia} suite, number of GPUs used during training, and the total number of GPU hours, calculated via (iteration time (s) $\times$ number of iterations $\times$ number of GPUs $\div$ 3600 s/hour). All GPUs are A100s with 40GB of memory.}
	\label{table:hardware}
\end{table}

Here ``total'' refers to training one model of each size in our suite. For this paper, we trained two models of each size (one on the Pile and one on the Pile deduplicated) and had to retrain both model suites an additional time as discussed in \cref{app:corrections}. Thus the total compute required for training the models for this paper was 544,280 A100-hours.

\clearpage

\section{Full Configuration Details}\label{app:hparam}

In \Cref{tab:config} we attach the full configuration details to train the models in this paper. Individual configuration files are available in the config files in our \href{https://github.com/EleutherAI/pythia}{GitHub Repository}.

\begin{table}[!ht]
\small
\centering
\begin{subtable}[h]{0.48\textwidth}
\begin{tabular}{lr}
    \toprule
    Configuration Key                               & Value \\
    \midrule
    attention-config & [[[``flash''], n-layers]] \\
    attention-dropout & 0 \\
    bias-gelu-fusion & True \\
    checkpoint-activations & True \\
    checkpoint-num-layers & 1 \\
    data-impl & mmap \\
    distributed-backend & nccl \\
    eval-interval & 143000 \\
    eval-iters & 10 \\
    fp16.enabled & True \\
    fp16.fp16 & True \\
    fp16.hysteresis & 2 \\
    fp16.initial-scale-power & 12 \\
    fp16.loss-scale & 0 \\
    fp16.loss-scale-window & 1000 \\
    fp16.min-loss-scale & 1 \\
    global-batch-size & 1024 \\
    gpt-j-residual & True \\
    gradient-accumulation-steps & -- \\
    gradient-clipping & 1.0 \\
    hidden-dropout & 0 \\
    hidden-size & -- \\
    init-method & small-init \\
    log-interval & 10 \\
    lr-decay-iters & 143000 \\
    lr-decay-style & cosine \\
    max-position-embeddings & 2048 \\
    min-lr & $0.1 * \text{optimizer.params.lr}$ \\
    model-parallel-size & -- \\
    no-weight-tying & True \\
    norm & layernorm \\
    num-attention-heads & -- \\\bottomrule
\end{tabular}
\end{subtable}
\begin{subtable}[h]{0.48\textwidth}
\begin{tabular}{lr}
    \toprule
    Configuration Key                               & Value \\
    \midrule
    num-layers & -- \\
    optimizer.params.betas & [0.9, 0.95] \\
    optimizer.params.eps & 1e-08 \\
    optimizer.params.lr & -- \\
    optimizer.type & Adam \\
    output-layer-init-method & wang-init \\
    output-layer-parallelism & column \\
    partition-activations & False \\
    pipe-parallel-size & 1 \\
    pos-emb & rotary \\
    rotary-pct & 0.25 \\
    save-interval & 1000 \\
    scaled-upper-triang-masked-softmax-fusion & True \\
    seq-length & 2048 \\
    split & 969,30,1\\
    steps-per-print & 10 \\
    synchronize-each-layer & True \\
    tokenizer-type & HFTokenizer \\
    train-iters & 143000 \\
    train-micro-batch-size-per-gpu & -- \\
    vocab-file & 20B-tokenizer.json \\
    wall-clock-breakdown & True \\
    warmup & 0.01 \\
    weight-decay & 0.01 \\
    zero-optimization.allgather-bucket-size & -- \\
    zero-optimization.allgather-partitions & True \\
    zero-optimization.contiguous-gradients & True \\
    zero-optimization.cpu-offload & False \\
    zero-optimization.overlap-comm & True \\
    zero-optimization.reduce-bucket-size & -- \\
    zero-optimization.reduce-scatter & True \\
    zero-optimization.stage & 1 \\\bottomrule
\end{tabular}
\end{subtable}
\caption{The full configuration details for \textit{Pythia} training. Exact model config files are also made available via our Github repository.}
\label{tab:config}
\end{table}

Configuration values marked with ``--'' differ between models. \Cref{table:interp} provides particular model dimensions. Additionally, some modifications are necessary to enable appropriate parallelism: while most models are trained with ``model-parallel-size = 1'', the 6.9b models were trained with ``model-parallel-size = 2'' and the 12b models were trained with ``model-parallel-size = 4''. Both these larger models were trained using ``zero-optimization.allgather-bucket-size = zero-optimization.reduce-bucket-size = 1260000000'', while all other models were trained with a value of 500000000. Exact number of GPUs, microbatch size per accelerator, and gradient accumulation steps per train step, for each model, are available in the config files in our Github repository.

\clearpage
\section{Additional Details on Design and Considerations}

\subsection{Assessment of Existing Suites}\label{app:suites}

We assessed existing model suites to determine if any pre-existing models met all of researchers' requirements and expectations for rigorous scientific study on language models. 

\paragraph{GPT-2 \citep{radford2019language}} No further notes.

\paragraph{GPT-3 \citep{brown2020language}} These models receive a half-mark for ``Public Models'' because while they have a publicly accessible API, the API costs money and OpenAI places substantial limitations on the research they allow you to do with the API. While these models are known to be similar to the models described in \citet{brown2020language}, they are not the same models. \citet{gao2021sizes} estimates the size of these models as being 350M, 1.3B, 6.7B, and 175B parameters respectively, which has been generally adopted by subsequent work. 

\paragraph{GPT-Neo \citep{black2021gpt,gpt-j,black2022gpt}} These models strictly speaking do not form a suite and have some non-negligible differences between them with respect to model architecture implementation, training codebase, tokenizer, and training data setup and order. Despite that, they are commmonly used \textit{as if they were} a consistent model suite.

\paragraph{OPT \citep{zhang2022opt}} While more checkpoints of OPT models exist (as is seen by their use in \citet{xia2022training}) they largely are not publicly available (less than 10 checkpoints available, only for the 2.7b, 6.7b, and 13b parameter models). Additionally, the training dataset for OPT is not public.

\paragraph{T5 \citep{raffel2020exploring}} The original paper did not release its training data, but it did release code for producing it which was subsequently run and released by \citet{dodge2021documenting}.

\paragraph{BLOOM \citep{scao2022bloom}} The ROOTS dataset that BLOOM was trained on is available via application to researchers, but the authors suggest that they may not make the full data indefinitely available in accompanying work \citep{jernite2022data,mcmillan2022documenting}. The BLOOM models were \textit{mostly} trained in a known and consistent order, however they handled training divergences by rewinding and skipping the offending sequences. Thus there are small (and undocumented) differences in the exact training composition and ordering across BLOOM models.

\subsection{Contrast with Multiply Trained Models}

A kind of dual question to the one considered in this paper regards how stable analysis of a particular model is when the random seed is allowed to vary. There are several model suites designed to answer this question, including the causal decoder Mistral suite \citep{karamcheti2021mistral} and the multiBERT suite \citep{sellam2021multiberts}. While we view this research as valuable, we ultimately decided against including several training runs of the same model in our suite because it would be ruinously expensive to do rigorously (doing 25 random seeds would cost approximately 10 million USD in compute) and we felt that the way to make the biggest impact with the resources we had available was to train one copy of each model.

\clearpage
\section{Evaluations}\label{app:evals}

We provide detailed evaluation scores and plots over the course of training for select benchmarks. In \cref{app:raw-zeroshot} and \cref{app:raw-fiveshot} we report raw scores for the final trained models, as well as comparisons to baseline model suites (\cref{app:baseline-evals}), on a number of standard NLP tasks, as well as scores for both model suites prior to the deduplicated Pythia models starting the second epoch on their training data, and in \cref{app:eval-plots} we provide plots of benchmarks over training. Full evaluation data, as well as evaluations on a wider range of tasks, can be found at \href{https://github.com/EleutherAI/pythia}{https://github.com/EleutherAI/pythia}.

\subsection{Raw Benchmark Scores - Zero Shot}\label{app:raw-zeroshot}
\begin{table}[H]
\centering
\begin{tabular}{rcccc}
\toprule
\textbf{Task} & \textbf{70M} & \textbf{160M} & \textbf{410M} & \textbf{1B} \\
\midrule
Lambada (OpenAI) & $0.185\pm0.005$ & $0.328\pm0.007$ & $0.516\pm0.007$ & $0.562\pm0.007$ \\
PIQA & $0.595\pm0.011$ & $0.627\pm0.011$ & $0.668\pm0.011$ & $0.707\pm0.011$ \\
WinoGrande & $0.528\pm0.014$ & $0.531\pm0.014$ & $0.537\pm0.014$ & $0.537\pm0.014$ \\
WSC & $0.365\pm0.047$ & $0.365\pm0.047$ & $0.567\pm0.049$ & $0.365\pm0.047$  \\
ARC - Easy & $0.374\pm0.010$ & $0.435\pm0.010$ & $0.521\pm0.010$ & $0.569\pm0.010$ \\
ARC - Challenge & $0.181\pm0.011$ & $0.188\pm0.011$ & $0.213\pm0.012$ & $0.244\pm0.013$  \\
SciQ & $0.601\pm0.015$ & $0.741\pm0.014$ & $0.811\pm0.012$ & $0.840\pm0.012$  \\
LogiQA & $0.210\pm0.016$ & $0.190\pm0.015$ & $0.220\pm0.016$ & $0.223\pm0.016$  \\
\bottomrule
\\
\textbf{Task} & \textbf{1.4B} & \textbf{2.8B} & \textbf{6.9B} & \textbf{12B} \\
\midrule
Lambada (OpenAI) & $0.616\pm0.007$ & $0.647\pm0.007$ & $0.673\pm0.007$ & $0.705\pm0.006$ \\
PIQA & $0.711\pm0.011$& $0.739\pm0.010$ & $0.752\pm0.010$ & $0.760\pm0.010$ \\
WinoGrande & $0.573\pm0.014$& $0.594\pm0.014$ & $0.609\pm0.014$ & $0.639\pm0.013$ \\
WSC & $0.365\pm0.047$ & $0.385\pm0.048$ & $0.365\pm0.047$ & $0.548\pm0.049$ \\
ARC - Easy & $0.606\pm0.010$ & $0.644\pm0.010$ & $0.673\pm0.010$ & $0.702\pm0.009$ \\
ARC - Challenge & $0.260\pm0.013$ & $0.295\pm0.013$ & $0.313\pm0.014$ & $0.318\pm0.014$ \\
SciQ & $0.865\pm0.011$ & $0.882\pm0.010$ & $0.897\pm0.010$ & $0.902\pm0.009$ \\
LogiQA & $0.210\pm0.016$& $0.212\pm0.016$ & $0.253\pm0.017$ & $0.224\pm0.016$ \\
\bottomrule
\end{tabular}
\caption{Zero-shot results on selected NLP Benchmarks, for the fully-trained Pythia suite.}
\end{table}

\begin{table}[H]
\centering
\begin{tabular}{rcccccccc}
\toprule
\textbf{Task} & \textbf{70M} & \textbf{160M} & \textbf{410M} & \textbf{1B} \\
\midrule
Lambada (OpenAI) & $0.192\pm0.005$ & $0.342\pm0.007$ & $0.524\pm0.007$ & $0.580\pm0.007$ \\
PIQA & $0.598\pm0.011$ & $0.618\pm0.011$ & $0.675\pm0.011$ & $0.700\pm0.011$ \\
WinoGrande & $0.492\pm0.014$ & $0.497\pm0.014$ & $0.534\pm0.014$ & $0.529\pm0.014$  \\
WSC & $0.365\pm0.047$ & $0.365\pm0.047$ & $0.471\pm0.049$ & $0.365\pm0.047$  \\
ARC - Easy & $0.385\pm0.010$ & $0.440\pm0.010$ & $0.517\pm0.010$ & $0.585\pm0.010$  \\
ARC - Challenge & $0.162\pm0.011$ & $0.201\pm0.012$ & $0.202\pm0.012$ & $0.245\pm0.013$  \\
SciQ & $0.606\pm0.015$ & $0.720\pm0.014$ & $0.826\pm0.012$ & $0.870\pm0.011$  \\
LogiQA & $0.235\pm0.017$ & $0.210\pm0.016$ & $0.209\pm0.016$ & $0.212\pm0.016$  \\
\bottomrule
\\
\textbf{Task} & \textbf{1.4B} & \textbf{2.8B} & \textbf{6.9B} & \textbf{12B} \\
\midrule
Lambada (OpenAI) & $0.619\pm0.007$& $0.651\pm0.007$ & $0.689\pm0.006$ & $0.710\pm0.006$ \\
PIQA & $0.720\pm0.010$ & $0.741\pm0.010$ & $0.760\pm0.010$ & $0.763\pm0.010$ \\
WinoGrande & $0.566\pm0.014$ & $0.582\pm0.014$ & $0.631\pm0.014$ & $0.660\pm0.013$ \\
WSC & $0.442\pm0.049$ & $0.385\pm0.048$ & $0.442\pm0.049$ & $0.394\pm0.048$ \\
ARC - Easy & $0.617\pm0.010$ & $0.635\pm0.010$ & $0.686\pm0.010$ & $0.708\pm0.009$ \\
ARC - Challenge & $0.272\pm0.013$ & $0.301\pm0.013$ & $0.331\pm0.014$ & $0.332\pm0.014$ \\
SciQ & $0.865\pm0.011$ & $0.882\pm0.010$ & $0.911\pm0.009$ & $0.929\pm0.008$ \\
LogiQA & $0.221\pm0.016$ & $0.214\pm0.016$ & $0.215\pm0.016$ & $0.224\pm0.016$ \\
\bottomrule
\end{tabular}
\caption{Zero-shot results on selected NLP Benchmarks, for the fully-trained Pythia (Deduplicated) suite.}
\end{table}

\begin{table}[H]
\centering
\begin{tabular}{rcccccccc}
\toprule
\textbf{Task} & \textbf{70M} & \textbf{160M} & \textbf{410M} & \textbf{1B} \\
\midrule
Lambada (OpenAI) & $0.214\pm0.006$ & $0.368\pm0.007$ & $0.500\pm0.007$ & $0.549\pm0.007$ \\
PIQA & $0.598\pm0.011$ & $0.625\pm0.011$ & $0.667\pm0.011$ & $0.701\pm0.011$ \\
WinoGrande & $0.508\pm0.014$ & $0.512\pm0.014$ & $0.525\pm0.014$ & $0.519\pm0.014$ \\
WSC & $0.365\pm0.047$ & $0.365\pm0.047$ & $0.625\pm0.048$ & $0.365\pm0.047$ \\
ARC - Easy & $0.359\pm0.010$ & $0.463\pm0.010$ & $0.512\pm0.010$ & $0.551\pm0.010$\\
ARC - Challenge & $0.172\pm0.011$ & $0.192\pm0.012$ & $0.218\pm0.012$ & $0.229\pm0.012$ \\
SciQ & $0.642\pm0.015$ & $0.764\pm0.013$ & $0.808\pm0.012$ & $0.837\pm0.012$ \\
LogiQA & $0.220\pm0.016$ & $0.214\pm0.016$ & $0.206\pm0.016$ & $0.224\pm0.016$ \\
\bottomrule
\\
\textbf{Task} & \textbf{1.4B} & \textbf{2.8B} & \textbf{6.9B} & \textbf{12B} \\
\midrule
Lambada (OpenAI) & $0.592\pm0.007$ & $0.633\pm0.007$ & $0.657\pm0.007$ & $0.684\pm0.006$ \\
PIQA & $0.705\pm0.011$ & $0.731\pm0.010$ & $0.741\pm0.010$ & $0.755\pm0.010$ \\
WinoGrande & $0.560\pm0.014$ & $0.592\pm0.014$ & $0.593\pm0.014$ & $0.630\pm0.014$ \\
WSC & $0.394\pm0.048$ & $0.365\pm0.047$ & $0.365\pm0.047$ & $0.635\pm0.047$ \\
ARC - Easy & $0.594\pm0.010$ & $0.622\pm0.010$ & $0.657\pm0.010$ & $0.686\pm0.010$ \\
ARC - Challenge & $0.253\pm0.013$ & $0.281\pm0.013$ & $0.318\pm0.014$ & $0.312\pm0.014$ \\
SciQ & $0.873\pm0.011$ & $0.875\pm0.010$ & $0.901\pm0.009$ & $0.909\pm0.009$ \\
LogiQA & $0.224\pm0.016$ & $0.220\pm0.016$ & $0.240\pm0.017$ & $0.230\pm0.017$ \\
\bottomrule
\end{tabular}
\caption{Zero-shot results on selected NLP Benchmarks, for the Pythia suite after 93k steps of pretraining (the closest step we measure prior to Pythia (Deduplicated) entering a second epoch at the 207B token mark).}
\end{table}

\begin{table}[H]
\centering
\begin{tabular}{rcccccccc}
\toprule
\textbf{Task} & \textbf{70M} & \textbf{160M} & \textbf{410M} & \textbf{1B} \\
\midrule
Lambada (OpenAI) & $0.230\pm0.006$ & $0.398\pm0.007$ & $0.529\pm0.007$ & $0.573\pm0.007$ \\
PIQA & $0.585\pm0.011$ & $0.628\pm0.011$ & $0.670\pm0.011$ & $0.696\pm0.011$ \\
WinoGrande & $0.511\pm0.014$ & $0.519\pm0.014$ & $0.530\pm0.014$ & $0.540\pm0.014$ \\
WSC & $0.365\pm0.047$ & $0.462\pm0.049$ & $0.625\pm0.048$ & $0.365\pm0.047$ \\
ARC - Easy & $0.380\pm0.010$ & $0.455\pm0.010$ & $0.526\pm0.010$ & $0.564\pm0.010$\\
ARC - Challenge & $0.177\pm0.011$ & $0.200\pm0.012$ & $0.209\pm0.012$ & $0.247\pm0.013$ \\
SciQ & $0.654\pm0.015$ & $0.774\pm0.013$ & $0.824\pm0.012$ & $0.858\pm0.011$ \\
LogiQA & $0.232\pm0.017$ & $0.217\pm0.016$ & $0.230\pm0.017$ & $0.224\pm0.016$ \\
\bottomrule
\\
\textbf{Task} & \textbf{1.4B} & \textbf{2.8B} & \textbf{6.9B} & \textbf{12B} \\
\midrule
Lambada (OpenAI) & $0.598\pm0.007$ & $0.633\pm0.007$ & $0.670\pm0.007$ & $0.697\pm0.006$ \\
PIQA & $0.715\pm0.011$ & $0.733\pm0.010$ & $0.746\pm0.010$ & $0.755\pm0.010$ \\
WinoGrande & $0.554\pm0.014$ & $0.583\pm0.014$ & $0.624\pm0.014$ & $0.636\pm0.014$ \\
WSC & $0.413\pm0.049$ & $0.365\pm0.047$ & $0.365\pm0.047$ & $0.500\pm0.049$ \\
ARC - Easy & $0.609\pm0.010$ & $0.622\pm0.010$ & $0.667\pm0.010$ & $0.691\pm0.009$ \\
ARC - Challenge & $0.266\pm0.013$ & $0.288\pm0.013$ & $0.319\pm0.014$ & $0.325\pm0.014$ \\
SciQ & $0.869\pm0.011$ & $0.882\pm0.010$ & $0.896\pm0.010$ & $0.925\pm0.008$ \\
LogiQA & $0.214\pm0.016$ & $0.209\pm0.016$ & $0.227\pm0.016$ & $0.220\pm0.016$ \\
\bottomrule
\end{tabular}
\caption{Zero-shot results on selected NLP Benchmarks, for the Pythia (Deduplicated) suite after 93k steps of pretraining (the closest step we measure prior to Pythia (Deduplicated) entering a second epoch at the 207B token mark).}
\end{table}

\subsection{Raw Benchmark Scores - Five Shot}\label{app:raw-fiveshot}

\begin{table}[H]
\centering
\begin{tabular}{rcccccccc}
\toprule
\textbf{Task} & \textbf{70M} & \textbf{160M} & \textbf{410M} & \textbf{1B} \\
\midrule
Lambada (OpenAI) & $0.125\pm0.005$ & $0.257\pm0.006$ & $0.455\pm0.007$ & $0.507\pm0.007$ \\
PIQA & $0.573\pm0.012$ & $0.621\pm0.011$ & $0.678\pm0.011$ & $0.705\pm0.011$ \\
WinoGrande & $0.522\pm0.014$ & $0.507\pm0.014$ & $0.530\pm0.014$ & $0.532\pm0.014$ \\
WSC & $0.365\pm0.047$ & $0.365\pm0.047$ & $0.365\pm0.047$ & $0.365\pm0.047$ \\
ARC - Easy & $0.381\pm0.010$ & $0.449\pm0.010$ & $0.555\pm0.010$ & $0.594\pm0.010$ \\
ARC - Challenge & $0.180\pm0.011$ & $0.186\pm0.011$ & $0.221\pm0.012$ & $0.259\pm0.013$ \\
SciQ & $0.577\pm0.016$ & $0.779\pm0.013$ & $0.891\pm0.010$ & $0.920\pm0.009$ \\
LogiQA & $0.218\pm0.016$ & $0.217\pm0.016$ & $0.220\pm0.016$ & $0.227\pm0.016$ \\
\bottomrule
\\
\textbf{Task} & \textbf{1.4B} & \textbf{2.8B} & \textbf{6.9B} & \textbf{12B} \\
\midrule
Lambada (OpenAI) & $0.578\pm0.007$ & $0.605\pm0.007$ & $0.638\pm0.007$ & $0.673\pm0.007$ \\
PIQA & $0.705\pm0.011$ & $0.736\pm0.010$ & $0.755\pm0.010$ & $0.760\pm0.010$ \\
WinoGrande & $0.580\pm0.014$ & $0.606\pm0.014$ & $0.637\pm0.014$ & $0.642\pm0.013$ \\
WSC & $0.365\pm0.047$ & $0.365\pm0.047$ & $0.365\pm0.047$ & $0.365\pm0.047$ \\
ARC - Easy & $0.643\pm0.010$& $0.673\pm0.010$ & $0.702\pm0.009$ & $0.710\pm0.009$ \\
ARC - Challenge & $0.290\pm0.013$& $0.323\pm0.014$ & $0.356\pm0.014$ & $0.365\pm0.014$ \\
SciQ & $0.92\pm0.009$ & $0.943\pm0.007$ & $0.951\pm0.007$ & $0.953\pm0.007$ \\
LogiQA & $0.240\pm0.017$ & $0.217\pm0.016$ & $0.270\pm0.017$ & $0.218\pm0.016$ \\
\bottomrule
\end{tabular}
\caption{Five-shot results on selected NLP Benchmarks, for the fully-trained Pythia suite.}
\end{table}

\begin{table}[H]
\centering
\begin{tabular}{rcccccccc}
\toprule
\textbf{Task} & \textbf{70M} & \textbf{160M} & \textbf{410M} & \textbf{1B} \\
\midrule
Lambada (OpenAI) & $0.134\pm0.005$ & $0.268\pm0.006$ & $0.466\pm0.007$ & $0.528\pm0.007$ \\
PIQA & $0.582\pm0.012$ & $0.620\pm0.011$ & $0.676\pm0.011$ & $0.704\pm0.011$ \\
WinoGrande & $0.499\pm0.014$ & $0.513\pm0.014$ & $0.536\pm0.014$ & $0.540\pm0.014$ \\
WSC & $0.365\pm0.047$ & $0.365\pm0.047$ & $0.365\pm0.047$ & $0.365\pm0.047$ \\
ARC - Easy & $0.383\pm0.010$ & $0.453\pm0.010$ & $0.539\pm0.010$ & $0.601\pm0.010$ \\
ARC - Challenge & $0.177\pm0.011$ & $0.205\pm0.012$ & $0.230\pm0.012$ & $0.260\pm0.013$ \\
SciQ & $0.598\pm0.016$ & $0.792\pm0.013$ & $0.880\pm0.010$ & $0.916\pm0.009$ \\
LogiQA & $0.250\pm0.017$ & $0.237\pm0.017$ & $0.210\pm0.016$ & $0.226\pm0.016$ \\
\bottomrule
\\
\textbf{Task} & \textbf{1.4B} & \textbf{2.8B} & \textbf{6.9B} & \textbf{12B} \\
\midrule
Lambada (OpenAI)& $0.568\pm0.007$ & $0.606\pm0.007$ & $0.663\pm0.007$ & $0.691\pm0.006$ \\
PIQA & $0.725\pm0.010$ & $0.734\pm0.010$ & $0.758\pm0.010$ & $0.767\pm0.010$ \\
WinoGrande & $0.569\pm0.014$ & $0.604\pm0.014$ & $0.638\pm0.014$ & $0.666\pm0.013$ \\
WSC & $0.365\pm0.047$ & $0.365\pm0.047$ & $0.365\pm0.047$ & $0.365\pm0.047$ \\
ARC - Easy & $0.633\pm0.001$ & $0.675\pm0.010$ & $0.702\pm0.009$ & $0.715\pm0.009$ \\
ARC - Challenge & $0.276\pm0.013$ & $0.329\pm0.014$ & $0.356\pm0.014$ & $0.368\pm0.014$ \\
SciQ & $0.926\pm0.008$ & $0.942\pm0.007$ & $0.952\pm0.007$ & $0.955\pm0.007$ \\
LogiQA & $0.230\pm0.017$ & $0.220\pm0.016$ & $0.257\pm0.017$ & $0.244\pm0.017$ \\
\bottomrule
\end{tabular}
\caption{Five-shot results on selected NLP Benchmarks, for the fully-trained Pythia (Deduplicated) suite.}
\end{table}

\begin{table}[H]
\centering
\begin{tabular}{rcccccccc}
\toprule
\textbf{Task} & \textbf{70M} & \textbf{160M} & \textbf{410M} & \textbf{1B} \\
\midrule
Lambada (OpenAI) & $0.134\pm0.005$ & $0.293\pm0.006$ & $0.433\pm0.007$ & $0.493\pm0.007$ \\
PIQA & $0.592\pm0.011$ & $0.627\pm0.011$ & $0.674\pm0.011$ & $0.693\pm0.011$ \\
WinoGrande & $0.531\pm0.014$ & $0.508\pm0.014$ & $0.530\pm0.014$ & $0.545\pm0.014$ \\
WSC & $0.365\pm0.047$ & $0.365\pm0.047$ & $0.365\pm0.047$ & $0.365\pm0.047$ \\
ARC - Easy & $0.375\pm0.010$ & $0.461\pm0.010$ & $0.544\pm0.010$ & $0.587\pm0.010$ \\
ARC - Challenge & $0.178\pm0.011$ & $0.194\pm0.012$ & $0.211\pm0.012$ & $0.261\pm0.013$ \\
SciQ & $0.605\pm0.015$ & $0.810\pm0.012$ & $0.889\pm0.010$ & $0.907\pm0.009$ \\
LogiQA & $0.223\pm0.016$ & $0.215\pm0.016$ & $0.229\pm0.016$ & $0.224\pm0.016$ \\
\bottomrule
\\
\textbf{Task} & \textbf{1.4B} & \textbf{2.8B} & \textbf{6.9B} & \textbf{12B} \\
\hline
Lambada (OpenAI) & $0.555\pm0.007$ & $0.590\pm0.007$ & $0.619\pm0.007$ & $0.650\pm0.007$ \\
PIQA & $0.697\pm0.011$ & $0.731\pm0.010$ & $0.748\pm0.010$ & $0.757\pm0.010$ \\
WinoGrande & $0.575\pm0.014$ & $0.603\pm0.014$ & $0.627\pm0.014$ & $0.639\pm0.014$ \\
WSC & $0.365\pm0.047$ & $0.365\pm0.047$ & $0.365\pm0.047$ & $0.356\pm0.047$ \\
ARC - Easy & $0.622\pm0.010$ & $0.667\pm0.010$ & $0.685\pm0.010$ & $0.702\pm0.009$ \\
ARC - Challenge & $0.283\pm0.013$ & $0.311\pm0.014$ & $0.351\pm0.014$ & $0.347\pm0.014$ \\
SciQ & $0.921\pm0.009$ & $0.942\pm0.007$ & $0.942\pm0.007$ & $0.952\pm0.007$ \\
LogiQA & $0.223\pm0.016$ & $0.215\pm0.016$ & $0.250\pm0.017$ & $0.229\pm0.016$ \\
\hline
\end{tabular}
\caption{Five-shot results on selected NLP Benchmarks, for the Pythia suite after 93k steps of pretraining (the closest step we measure prior to Pythia (Deduplicated) entering a second epoch at the 207B token mark).}
\end{table}

\begin{table}[H]
\centering
\begin{tabular}{rcccccccc}
\toprule
\textbf{Task} & \textbf{70M} & \textbf{160M} & \textbf{410M} & \textbf{1B} \\
\midrule
Lambada (OpenAI) & $0.153\pm0.005$ & $0.333\pm0.007$ & $0.468\pm0.007$ & $0.513\pm0.007$ \\
PIQA & $0.589\pm0.011$ & $0.628\pm0.011$ & $0.671\pm0.011$ & $0.697\pm0.011$\\
WinoGrande & $0.515\pm0.014$ & $0.513\pm0.014$ & $0.542\pm0.014$ & $0.558\pm0.014$ \\
WSC & $0.365\pm0.047$ & $0.365\pm0.047$ & $0.365\pm0.047$ & $0.365\pm0.047$ \\
ARC - Easy & $0.392\pm0.010$ & $0.468\pm0.010$ & $0.540\pm0.010$ & $0.593\pm0.010$ \\
ARC — Challenge & $0.172\pm0.011$ & $0.201\pm0.012$ & $0.231\pm0.012$ & $0.250\pm0.013$ \\
SciQ & $0.600\pm0.015$ & $0.815\pm0.012$ & $0.877\pm0.010$ & $0.913\pm0.009$ \\
LogiQA & $0.238\pm0.017$ & $0.238\pm0.017$ & $0.209\pm0.016$ & $0.214\pm0.016$ \\
\bottomrule
\\
\textbf{Task} & \textbf{1.4B} & \textbf{2.8B} & \textbf{6.9B} & \textbf{12B} \\
\midrule
Lambada (OpenAI) & $0.563\pm0.007$ & $0.593\pm0.007$ & $0.652\pm0.007$ & $0.685\pm0.006$ \\
PIQA & $0.712\pm0.011$ & $0.727\pm0.010$ & $0.750\pm0.010$ & $0.751\pm0.010$ \\
WinoGrande & $0.567\pm0.014$ & $0.596\pm0.014$ & $0.636\pm0.014$ & $0.643\pm0.013$ \\
WSC & $0.365\pm0.047$ & $0.365\pm0.047$ & $0.346\pm0.047$ & $0.365\pm0.047$ \\
ARC - Easy & $0.630\pm0.010$ & $0.664\pm0.010$ & $0.683\pm0.010$ & $0.712\pm0.009$ \\
ARC — Challenge & $0.274\pm0.013$ & $0.310\pm0.014$ & $0.355\pm0.014$ & $0.369\pm0.014$ \\
SciQ & $0.918\pm0.009$ & $0.942\pm0.007$ & $0.947\pm0.007$ & $0.948\pm0.007$ \\
LogiQA & $0.229\pm0.017$ & $0.220\pm0.016$ & $0.240\pm0.017$ & $0.229\pm0.016$ \\
\bottomrule
\end{tabular}
\caption{Five-shot results on selected NLP Benchmarks, for the Pythia (Deduplicated) suite after 93k steps of pretraining (the closest step we measure prior to Pythia (Deduplicated) entering a second epoch at the 207B token mark).}
\end{table}

\subsection{Comparison to Baseline Models}\label{app:baseline-evals}
\begin{table}[H]
\centering
\begin{tabular}{rccccc}
\toprule
\textbf{Task} & \textbf{560M} & \textbf{1.1B} & \textbf{1.7B} & \textbf{3B} \\
\midrule
Lambada (OpenAI) & $0.341\pm0.007$ & $0.426\pm0.007$ & $0.462\pm0.007$ & $0.518\pm0.007$ \\
PIQA & $0.637\pm0.011$ & $0.672\pm0.011$ & $0.686\pm0.011$ & $0.708\pm0.011$ \\
WinoGrande & $0.504\pm0.014$ & $0.547\pm0.014$ & $0.572\pm0.014$ & $0.586\pm0.014$ \\
WSC & $0.442\pm0.049$ & $0.365\pm0.047$ & $0.365\pm0.047$ & $0.375\pm0.048$ \\
ARC - Easy & $0.476\pm0.010$ & $0.515\pm0.010$ & $0.562\pm0.010$ & $0.594\pm0.010$ \\
ARC — Challenge & $0.221\pm0.012$ & $0.236\pm0.012$ & $0.238\pm0.012$ & $0.280\pm0.013$ \\
SciQ & $0.804\pm0.013$ & $0.833\pm0.012$ & $0.851\pm0.011$ & $0.891\pm0.010$ \\
LogiQA & $0.217\pm0.016$ & $0.189\pm0.015$ & $0.217\pm0.016$ & $0.206\pm0.016$ \\
\bottomrule
\end{tabular}
\caption{Zero-shot results on standard NLP benchmarks for the BLOOM model suite, reported for comparison with Pythia's performance.}
\end{table}

\begin{table}[H]
\centering
\begin{tabular}{rccccc}
\hline
\textbf{Task} & \textbf{125M} & \textbf{350M} & \textbf{1.3B} & \textbf{2.7B} \\
\hline
Lambada (OpenAI) & $0.379\pm0.007$ & $0.452\pm0.007$ & $0.579\pm0.007$ & $0.636\pm0.007$ \\
PIQA & $0.630\pm0.011$ & $0.644\pm0.011$ & $0.717\pm0.011$ & $0.739\pm0.010$ \\
WinoGrande & $0.503\pm0.014$ & $0.523\pm0.014$ & $0.597\pm0.014$ & $0.610\pm0.014$ \\
WSC & $0.365\pm0.047$ & $0.365\pm0.047$ & $0.385\pm0.048$ & $0.635\pm0.047$ \\
ARC - Easy & $0.435\pm0.010$ & $0.440\pm0.010$ & $0.570\pm0.010$ & $0.608\pm0.010$ \\
ARC — Challenge & $0.189\pm0.011$ & $0.207\pm0.012$ & $0.231\pm0.012$ & $0.268\pm0.013$ \\
SciQ & $0.751\pm0.014$ & $0.748\pm0.014$ & $0.845\pm0.011$ & $0.858\pm0.011$ \\
LogiQA & $0.227\pm0.016$ & $0.210\pm0.016$ & $0.223\pm0.016$ & $0.210\pm0.016$ \\
\bottomrule
\\
\textbf{Task} & \textbf{6.7B} & \textbf{13B} & \textbf{30B} & \textbf{66B} \\
\midrule
Lambada (OpenAI) & $0.677\pm0.007$ & $0.686\pm0.006$ & $0.715\pm0.006$ & $0.739\pm0.006$ \\
PIQA & $0.763\pm0.010$ & $0.760\pm0.010$ & $0.776\pm0.010$ & $0.788\pm0.010$ \\
WinoGrande & $0.653\pm0.013$ & $0.652\pm0.013$ & $0.682\pm0.013$ & $0.687\pm0.013$ \\
WSC & $0.423\pm0.049$ & $0.606\pm0.048$ & $0.596\pm0.048$ & $0.548\pm0.049$ \\
ARC - Easy & $0.656\pm0.010$ & $0.671\pm0.010$ & $0.700\pm0.009$ & $0.717\pm0.009$ \\
ARC — Challenge & $0.305\pm0.013$ & $0.329\pm0.014$ & $0.346\pm0.014$ & $0.372\pm0.014$ \\
SciQ & $0.901\pm0.009$ & $0.908\pm0.009$ & $0.911\pm0.009$ & $0.926\pm0.008$ \\
LogiQA & $0.235\pm0.017$ & $0.227\pm0.016$ & $0.217\pm0.016$ & $0.227\pm0.016$ \\
\bottomrule
\end{tabular}
\caption{Zero-shot results on standard NLP benchmarks for the OPT model suite up to 66B parameters, reported for comparison with Pythia's performance.}
\end{table}

\clearpage

\subsection{Graphs}\label{app:eval-plots}
\usepgfplotslibrary{fillbetween}
\usetikzlibrary {decorations}
\usepgfplotslibrary{external}
\pgfplotsset{compat=1.16}
\usetikzlibrary{external}\tikzexternalize

\def\evalplot#1{\begin{tikzpicture}
\begin{axis}[small, width=0.98\columnwidth,  legend cell align={left}, legend pos={south east}, legend columns={1}, legend style={font=\scriptsize}, anchor={north west}, xshift={1.0mm}, yshift={-1.0mm}, scaled x ticks={false}, xlabel={Size (Parameters)}, xmode={log}, log basis x={10}, xmajorgrids={true}, xmin={1.0e7}, xmax={1.0e11}, xticklabels={{$10^{7}$,$10^{8}$,$10^{9}$,$10^{10}$,$10^{11}$}}, xtick={{1.0e7,1.0e8,1.0e9,1.0e10,1.0e11}}, xtick align={inside}, x grid style={draw opacity={0.1}, line width={0.5}, solid}, extra x ticks={{2.0e7,3.0e7,4.0e7,5.0e7,6.0e7,7.0e7,8.0e7,9.0e7,2.0e8,3.0e8,4.0e8,5.0e8,6.0e8,7.0e8,8.0e8,9.0e8,2.0e9,3.0e9,4.0e9,5.0e9,6.0e9,7.0e9,8.0e9,9.0e9,2.0e10,3.0e10,4.0e10,5.0e10,6.0e10,7.0e10,8.0e10,9.0e10}}, extra x tick labels={}, extra x tick style={grid={major}, major tick length={0.1cm}}, axis x line*={left}, scaled y ticks={false}, ylabel={Accuracy}, ymajorgrids={true}, ymin={0}, ymax={1}, yticklabels={{$0.0$,$0.2$,$0.4$,$0.6$,$0.8$,$1.0$}}, ytick={{0.0,0.2,0.4,0.6,0.8,1.0}}, ytick align={inside}, y grid style={draw opacity={0.1}, line width={0.5}, solid}, extra y ticks={{0.04,0.08,0.12,0.16,0.24,0.28,0.32,0.36,0.44,0.48,0.52,0.56,0.64,0.68,0.72,0.76,0.84,0.88,0.92,0.96}}, extra y tick labels={}, extra y tick style={grid={major}, major tick length={0.1cm}}, axis y line*={left}, colorbar={false}]
    \addplot[color={rgb,1:red,0.0039;green,0.451;blue,0.698}, solid, error bars/.cd, y explicit, y dir=both]
        table [
                x=size,
                y={#1-acc},
                y error={#1-acc_stderr},
            ] {appendices/evaluation_data/opt.dat}
        ;
    \addlegendentry {OPT}
    \addplot[color={rgb,1:red,0.8706;green,0.5608;blue,0.0196}, solid,  error bars/.cd, y explicit, y dir=both]
        table [
                x=size,
                y={#1-acc},
                y error={#1-acc_stderr},
            ] {appendices/evaluation_data/bloom.dat}
        ;
    \addlegendentry {BLOOM}
    \addplot[color={rgb,1:red,0.0078;green,0.6196;blue,0.451}, solid,  x filter/.append expression={\thisrow{num_shots} == 0 ? \pgfmathresult : NaN}, x filter/.append expression={\thisrow{tokens_seen} == 299892736000 ? \pgfmathresult : NaN}, error bars/.cd, y explicit, y dir=both]
        table [
                x=size,
                y={#1-acc},
                y error={#1-acc_stderr},
            ] {appendices/evaluation_data/pythia_standard.dat}
        ;
    \addlegendentry {Pythia}
    \addplot[color={rgb,1:red,0.835;green,0.368;blue,0.000}, solid,  x filter/.append expression={\thisrow{num_shots} == 0 ? \pgfmathresult : NaN}, x filter/.append expression={\thisrow{tokens_seen} == 299892736000 ? \pgfmathresult : NaN}, error bars/.cd, y explicit, y dir=both]
        table [
                x=size,
                y={#1-acc},
                y error={#1-acc_stderr},
            ] {appendices/evaluation_data/pythia_deduped.dat}
        ;
    \addlegendentry {Pythia (Deduplicated)}
\end{axis}
\end{tikzpicture}}

\def\deltaevalplot#1{\begin{tikzpicture}
\begin{axis}[small, width=0.9\columnwidth, legend cell align={left}, legend pos={south east}, legend columns={1}, legend style={font=\scriptsize}, anchor={north west}, xshift={1.0mm}, yshift={-1.0mm}, scaled x ticks={false}, xlabel={Size (Parameters)}, xmode={log}, log basis x={10}, xmajorgrids={true}, xmin={1.0e7}, xmax={1.25e10}, xticklabels={{$10^{7}$,$10^{8}$,$10^{9}$,$10^{10}$}}, xtick={{1.0e7,1.0e8,1.0e9,1.0e10}}, xtick align={inside}, x grid style={draw opacity={0.1}, line width={0.5}, solid}, extra x ticks={{2.0e7,3.0e7,4.0e7,5.0e7,6.0e7,7.0e7,8.0e7,9.0e7,2.0e8,3.0e8,4.0e8,5.0e8,6.0e8,7.0e8,8.0e8,9.0e8,2.0e9,3.0e9,4.0e9,5.0e9,6.0e9,7.0e9,8.0e9,9.0e9}}, extra x tick labels={}, extra x tick style={grid={major}, major tick length={0.1cm}}, axis x line*={left}, scaled y ticks={false}, ylabel={Accuracy}, ymajorgrids={true}, ymin={0}, ymax={1}, yticklabels={{$0.0$,$0.2$,$0.4$,$0.6$,$0.8$,$1.0$}}, ytick={{0.0,0.2,0.4,0.6,0.8,1.0}}, ytick align={inside}, y grid style={draw opacity={0.1}, line width={0.5}, solid}, extra y ticks={{0.04,0.08,0.12,0.16,0.24,0.28,0.32,0.36,0.44,0.48,0.52,0.56,0.64,0.68,0.72,0.76,0.84,0.88,0.92,0.96}}, extra y tick labels={}, extra y tick style={grid={major}, major tick length={0.1cm}}, axis y line*={left}, colorbar={false}]
    \addplot[color={rgb,1:red,0.0078;green,0.6196;blue,0.451}, solid,  x filter/.append expression={\thisrow{num_shots} == 0 ? \pgfmathresult : NaN}, x filter/.append expression={\thisrow{tokens_seen} == 195035136000 ? \pgfmathresult : NaN}, error bars/.cd, y explicit, y dir=both]
        table [
                x=size,
                y={#1-acc},
                y error={#1-acc_stderr},
            ] {appendices/evaluation_data/pythia_standard.dat}
        ;
    \addlegendentry {Pythia}
    \addplot[color={rgb,1:red,0.835;green,0.368;blue,0.000}, solid,  x filter/.append expression={\thisrow{num_shots} == 0 ? \pgfmathresult : NaN}, x filter/.append expression={\thisrow{tokens_seen} == 195035136000 ? \pgfmathresult : NaN}, error bars/.cd, y explicit, y dir=both]
        table [
                x=size,
                y={#1-acc},
                y error={#1-acc_stderr},
            ] {appendices/evaluation_data/pythia_deduped.dat}
        ;
    \addlegendentry {Pythia (Deduplicated)}
\end{axis}
\end{tikzpicture}}

\def\evaltimeplot#1#2#3#4#5#6{\begin{tikzpicture}
\begin{axis}[small, width=0.9\columnwidth, height=0.9\columnwidth, legend cell align={left}, legend pos={#5}, legend columns={#6}, legend style={font=\scriptsize}, anchor={north west}, xshift={1.0mm}, yshift={-1.0mm}, scaled x ticks={false}, xlabel={Tokens}, xmode={log}, log basis x={10}, xmajorgrids={true}, xmin={1.0e6}, xmax={4.0e11}, xticklabels={{$10^{6}$,$10^{7}$,$10^{8}$,$10^{9}$,$10^{10}$,$10^{11}$}}, xtick={{1.0e6,1.0e7,1.0e8,1.0e9,1.0e10,1.0e11}}, xtick align={inside}, x grid style={draw opacity={0.1}, line width={0.5}, solid}, extra x ticks={{2.0e6,3.0e6,4.0e6,5.0e6,6.0e6,7.0e6,8.0e6,9.0e6,2.0e7,3.0e7,4.0e7,5.0e7,6.0e7,7.0e7,8.0e7,9.0e7,2.0e8,3.0e8,4.0e8,5.0e8,6.0e8,7.0e8,8.0e8,9.0e8,2.0e9,3.0e9,4.0e9,5.0e9,6.0e9,7.0e9,8.0e9,9.0e9,2.0e10,3.0e10,4.0e10,5.0e10,6.0e10,7.0e10,8.0e10,9.0e10,2.0e11,3.0e11, 4.0e11}}, extra x tick labels={}, extra x tick style={grid={major}, major tick length={0.1cm}}, axis x line*={left}, scaled y ticks={false}, ylabel={Accuracy}, ymajorgrids={true}, ymin={0}, ymax={1}, yticklabels={{$0.0$,$0.2$,$0.4$,$0.6$,$0.8$,$1.0$}}, ytick={{0.0,0.2,0.4,0.6,0.8,1.0}}, ytick align={inside}, y grid style={draw opacity={0.1}, line width={0.5}, solid}, extra y ticks={{0.04,0.08,0.12,0.16,0.24,0.28,0.32,0.36,0.44,0.48,0.52,0.56,0.64,0.68,0.72,0.76,0.84,0.88,0.92,0.96}}, extra y tick labels={}, extra y tick style={grid={major}, major tick length={0.1cm}}, axis y line*={left}, colorbar={false}]
    \addplot[color={rgb,1:red,0.0039;green,0.451;blue,0.698}, solid, x filter/.append expression={\thisrow{num_shots} == #3 ? \pgfmathresult : NaN}, x filter/.append expression={\thisrow{size} == 18874368 ? \pgfmathresult : NaN},  error bars/.cd, y explicit, y dir=both]
        table [
                x={tokens_seen},
                y={#1-#2},
                y error={#1-#2_stderr},
            ] {appendices/evaluation_data/pythia_#4.dat}
        ;
    \addlegendentry {70M}
    \addplot[color={rgb,1:red,0.8706;green,0.5608;blue,0.0196}, solid, x filter/.append expression={\thisrow{num_shots} == #3 ? \pgfmathresult : NaN}, x filter/.append expression={\thisrow{size} == 84934656 ? \pgfmathresult : NaN},  error bars/.cd, y explicit, y dir=both]
        table [
                x={tokens_seen},
                y={#1-#2},
                y error={#1-#2_stderr},
            ] {appendices/evaluation_data/pythia_#4.dat}
        ;
    \addlegendentry {160M}
    \addplot[color={rgb,1:red,0.0078;green,0.6196;blue,0.451}, solid, x filter/.append expression={\thisrow{num_shots} == #3 ? \pgfmathresult : NaN}, x filter/.append expression={\thisrow{size} == 301989888 ? \pgfmathresult : NaN},  error bars/.cd, y explicit, y dir=both]
        table [
                x={tokens_seen},
                y={#1-#2},
                y error={#1-#2_stderr},
            ] {appendices/evaluation_data/pythia_#4.dat}
        ;
    \addlegendentry {410M}
    \addplot[color={rgb,1:red,0.835;green,0.368;blue,0.000}, solid, x filter/.append expression={\thisrow{num_shots} == #3 ? \pgfmathresult : NaN}, x filter/.append expression={\thisrow{size} == 805306368 ? \pgfmathresult : NaN},  error bars/.cd, y explicit, y dir=both]
        table [
                x={tokens_seen},
                y={#1-#2},
                y error={#1-#2_stderr},
            ] {appendices/evaluation_data/pythia_#4.dat}
        ;
    \addlegendentry {1.0B}
    \addplot[color={rgb,1:red,0.8;green,0.471;blue,0.7373}, solid, x filter/.append expression={\thisrow{num_shots} == #3 ? \pgfmathresult : NaN}, x filter/.append expression={\thisrow{size} == 1207959552 ? \pgfmathresult : NaN},  error bars/.cd, y explicit, y dir=both]
        table [
                x={tokens_seen},
                y={#1-#2},
                y error={#1-#2_stderr},
            ] {appendices/evaluation_data/pythia_#4.dat}
        ;
    \addlegendentry {1.4B}
    \addplot[color={rgb,1:red,0.792;green,0.5686;blue,0.3803}, solid, x filter/.append expression={\thisrow{num_shots} == #3 ? \pgfmathresult : NaN}, x filter/.append expression={\thisrow{size} == 2516582400 ? \pgfmathresult : NaN},  error bars/.cd, y explicit, y dir=both]
        table [
                x={tokens_seen},
                y={#1-#2},
                y error={#1-#2_stderr},
            ] {appendices/evaluation_data/pythia_#4.dat}
        ;
    \addlegendentry {2.8B}
    \addplot[color={rgb,1:red,0.9843;green,0.6863;blue,0.8941}, solid, x filter/.append expression={\thisrow{num_shots} == #3 ? \pgfmathresult : NaN}, x filter/.append expression={\thisrow{size} == 6442450944 ? \pgfmathresult : NaN},  error bars/.cd, y explicit, y dir=both]
        table [
                x={tokens_seen},
                y={#1-#2},
                y error={#1-#2_stderr},
            ] {appendices/evaluation_data/pythia_#4.dat}
        ;
    \addlegendentry {6.9B}
    \addplot[color={rgb,1:red,0.5804;green,0.5804;blue,0.5804}, solid, x filter/.append expression={\thisrow{num_shots} == #3 ? \pgfmathresult : NaN}, x filter/.append expression={\thisrow{size} == 11324620800 ? \pgfmathresult : NaN},  error bars/.cd, y explicit, y dir=both]
        table [
                x={tokens_seen},
                y={#1-#2},
                y error={#1-#2_stderr},
            ] {appendices/evaluation_data/pythia_#4.dat}
        ;
    \addlegendentry {12B}
\end{axis}
\end{tikzpicture}}

\def\evaltimeplotepoch#1#2#3#4#5#6{\begin{tikzpicture}
\begin{axis}[small, width=0.9\columnwidth,height=0.9\columnwidth, legend cell align={left}, legend pos={#5}, legend columns={#6}, legend style={font=\scriptsize}, anchor={north west}, xshift={1.0mm}, yshift={-1.0mm}, scaled x ticks={false}, xlabel={Tokens}, xmode={log}, log basis x={10}, xmajorgrids={true}, xmin={1.0e6}, xmax={4.0e11}, xticklabels={{$10^{6}$,$10^{7}$,$10^{8}$,$10^{9}$,$10^{10}$,$10^{11}$}}, xtick={{1.0e6,1.0e7,1.0e8,1.0e9,1.0e10,1.0e11}}, xtick align={inside}, x grid style={draw opacity={0.1}, line width={0.5}, solid}, extra x ticks={{2.0e6,3.0e6,4.0e6,5.0e6,6.0e6,7.0e6,8.0e6,9.0e6,2.0e7,3.0e7,4.0e7,5.0e7,6.0e7,7.0e7,8.0e7,9.0e7,2.0e8,3.0e8,4.0e8,5.0e8,6.0e8,7.0e8,8.0e8,9.0e8,2.0e9,3.0e9,4.0e9,5.0e9,6.0e9,7.0e9,8.0e9,9.0e9,2.0e10,3.0e10,4.0e10,5.0e10,6.0e10,7.0e10,8.0e10,9.0e10,2.0e11,3.0e11, 4.0e11}}, extra x tick labels={}, extra x tick style={grid={major}, major tick length={0.1cm}}, axis x line*={left}, scaled y ticks={false}, ylabel={Accuracy}, ymajorgrids={true}, ymin={0}, ymax={1}, yticklabels={{$0.0$,$0.2$,$0.4$,$0.6$,$0.8$,$1.0$}}, ytick={{0.0,0.2,0.4,0.6,0.8,1.0}}, ytick align={inside}, y grid style={draw opacity={0.1}, line width={0.5}, solid}, extra y ticks={{0.04,0.08,0.12,0.16,0.24,0.28,0.32,0.36,0.44,0.48,0.52,0.56,0.64,0.68,0.72,0.76,0.84,0.88,0.92,0.96}}, extra y tick labels={}, extra y tick style={grid={major}, major tick length={0.1cm}}, axis y line*={left}, colorbar={false}]
    \addplot[color={rgb,1:red,0.0039;green,0.451;blue,0.698}, solid, x filter/.append expression={\thisrow{num_shots} == #3 ? \pgfmathresult : NaN}, x filter/.append expression={\thisrow{size} == 18874368 ? \pgfmathresult : NaN},  error bars/.cd, y explicit, y dir=both]
        table [
                x={tokens_seen},
                y={#1-#2},
                y error={#1-#2_stderr},
            ] {appendices/evaluation_data/pythia_#4.dat}
        ;
    \addlegendentry {70M}
    \addplot[color={rgb,1:red,0.8706;green,0.5608;blue,0.0196}, solid, x filter/.append expression={\thisrow{num_shots} == #3 ? \pgfmathresult : NaN}, x filter/.append expression={\thisrow{size} == 84934656 ? \pgfmathresult : NaN},  error bars/.cd, y explicit, y dir=both]
        table [
                x={tokens_seen},
                y={#1-#2},
                y error={#1-#2_stderr},
            ] {appendices/evaluation_data/pythia_#4.dat}
        ;
    \addlegendentry {160M}
    \addplot[color={rgb,1:red,0.0078;green,0.6196;blue,0.451}, solid, x filter/.append expression={\thisrow{num_shots} == #3 ? \pgfmathresult : NaN}, x filter/.append expression={\thisrow{size} == 301989888 ? \pgfmathresult : NaN},  error bars/.cd, y explicit, y dir=both]
        table [
                x={tokens_seen},
                y={#1-#2},
                y error={#1-#2_stderr},
            ] {appendices/evaluation_data/pythia_#4.dat}
        ;
    \addlegendentry {410M}
    \addplot[color={rgb,1:red,0.835;green,0.368;blue,0.000}, solid, x filter/.append expression={\thisrow{num_shots} == #3 ? \pgfmathresult : NaN}, x filter/.append expression={\thisrow{size} == 805306368 ? \pgfmathresult : NaN},  error bars/.cd, y explicit, y dir=both]
        table [
                x={tokens_seen},
                y={#1-#2},
                y error={#1-#2_stderr},
            ] {appendices/evaluation_data/pythia_#4.dat}
        ;
    \addlegendentry {1.0B}
    \addplot[color={rgb,1:red,0.8;green,0.471;blue,0.7373}, solid, x filter/.append expression={\thisrow{num_shots} == #3 ? \pgfmathresult : NaN}, x filter/.append expression={\thisrow{size} == 1207959552 ? \pgfmathresult : NaN},  error bars/.cd, y explicit, y dir=both]
        table [
                x={tokens_seen},
                y={#1-#2},
                y error={#1-#2_stderr},
            ] {appendices/evaluation_data/pythia_#4.dat}
        ;
    \addlegendentry {1.4B}
    \addplot[color={rgb,1:red,0.792;green,0.5686;blue,0.3803}, solid, x filter/.append expression={\thisrow{num_shots} == #3 ? \pgfmathresult : NaN}, x filter/.append expression={\thisrow{size} == 2516582400 ? \pgfmathresult : NaN},  error bars/.cd, y explicit, y dir=both]
        table [
                x={tokens_seen},
                y={#1-#2},
                y error={#1-#2_stderr},
            ] {appendices/evaluation_data/pythia_#4.dat}
        ;
    \addlegendentry {2.8B}
    \addplot[color={rgb,1:red,0.9843;green,0.6863;blue,0.8941}, solid, x filter/.append expression={\thisrow{num_shots} == #3 ? \pgfmathresult : NaN}, x filter/.append expression={\thisrow{size} == 6442450944 ? \pgfmathresult : NaN},  error bars/.cd, y explicit, y dir=both]
        table [
                x={tokens_seen},
                y={#1-#2},
                y error={#1-#2_stderr},
            ] {appendices/evaluation_data/pythia_#4.dat}
        ;
    \addlegendentry {6.9B}
    \addplot[color={rgb,1:red,0.5804;green,0.5804;blue,0.5804}, solid, x filter/.append expression={\thisrow{num_shots} == #3 ? \pgfmathresult : NaN}, x filter/.append expression={\thisrow{size} == 11324620800 ? \pgfmathresult : NaN},  error bars/.cd, y explicit, y dir=both]
        table [
                x={tokens_seen},
                y={#1-#2},
                y error={#1-#2_stderr},
            ] {appendices/evaluation_data/pythia_#4.dat}
        ;
    \addlegendentry {12B}
    \draw [dashed, dash expand off, color=black!50!gray] (207e9,1) node[anchor=north east]{\scriptsize End of Epoch 1} -- (207e9,0);
\end{axis}
\end{tikzpicture}}

\begin{figure}[h!]
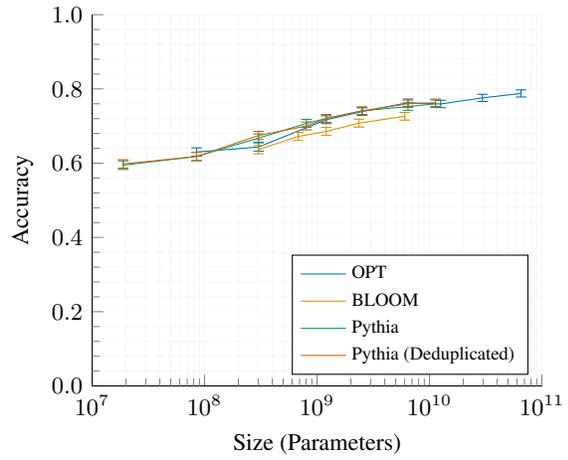
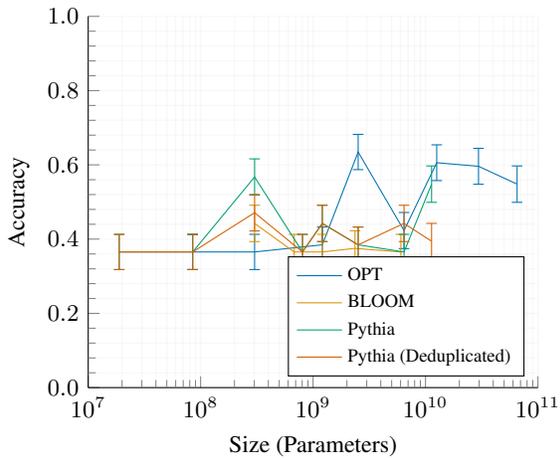
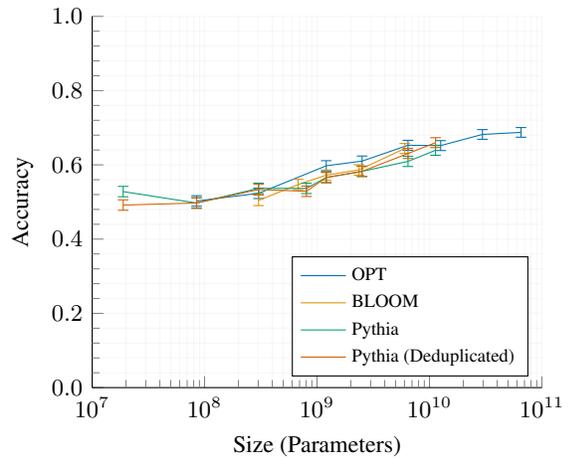
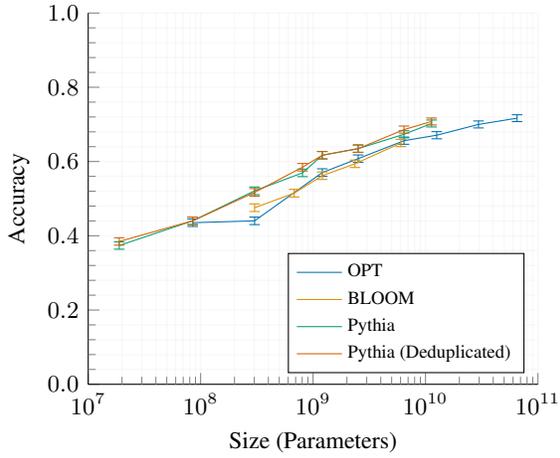
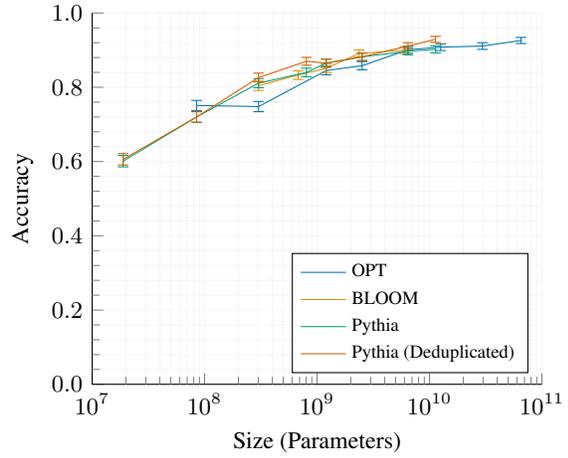

\centering
\begin{subfigure}{0.45\columnwidth}
\centering
\evalplot{lambada_openai}
\caption{LAMBADA (OpenAI)}
\end{subfigure}
\begin{subfigure}{0.45\columnwidth}
\centering
\evalplot{piqa}
\caption{PIQA}
\end{subfigure}
\begin{subfigure}{0.45\columnwidth}
\centering
\evalplot{wsc}
\caption{Winogrand Schema Challenge}
\end{subfigure}
\begin{subfigure}{0.45\columnwidth}
\centering
\evalplot{winogrande}
\caption{WinoGrande}
\end{subfigure}
\begin{subfigure}{0.45\columnwidth}
\centering
\evalplot{arc_easy}
\caption{AI2 Reasoning Challenge --- Easy Set}
\end{subfigure}
\begin{subfigure}{0.45\columnwidth}
\centering
\evalplot{sciq}
\caption{SciQ}
\end{subfigure}
\caption{Zero-shot evaluations of final Pythia checkpoints against OPT and BLOOM.}
\end{figure}

\begin{figure}[h]
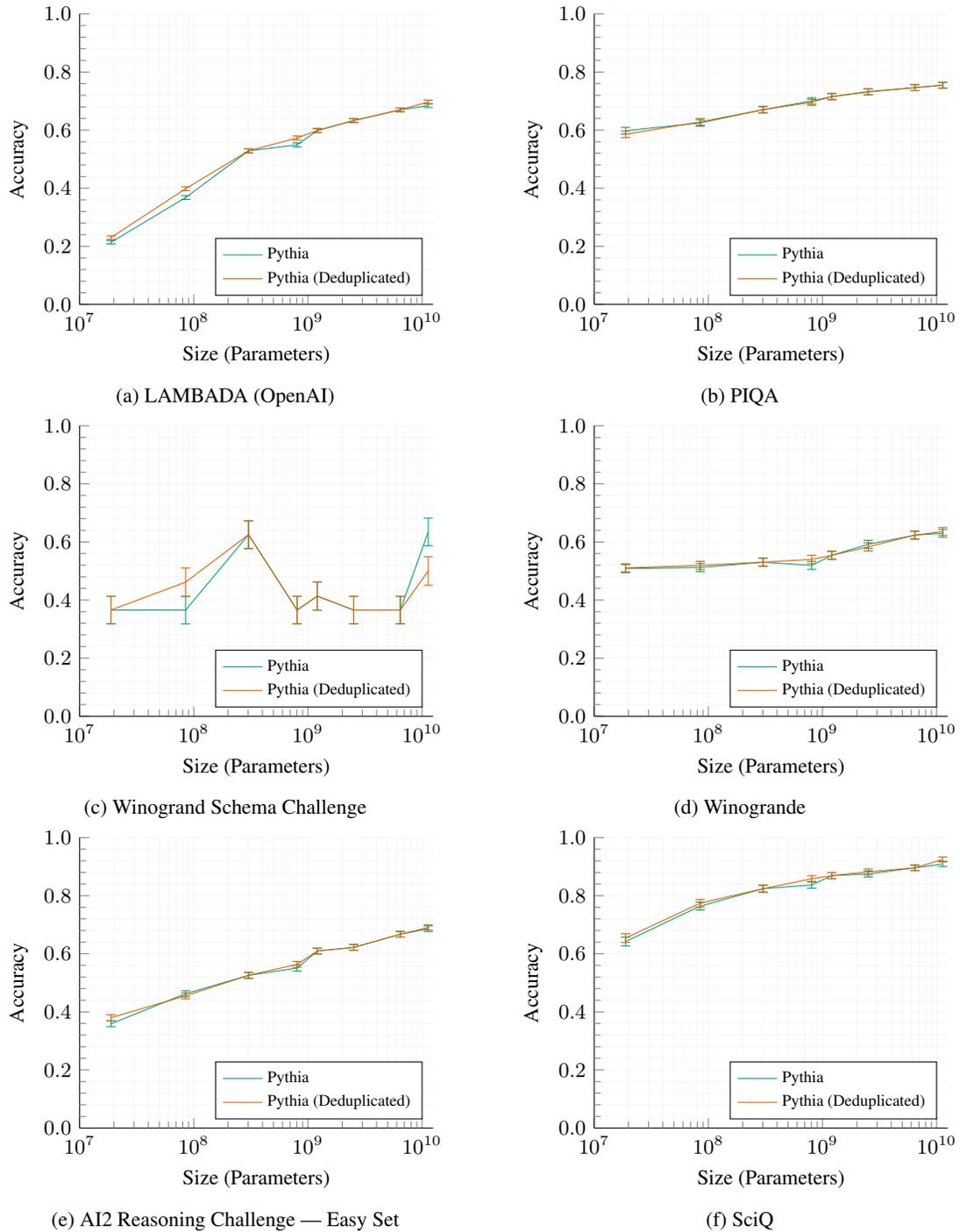

\centering
\begin{subfigure}{0.45\columnwidth}
\centering
\deltaevalplot{lambada_openai}
\caption{LAMBADA (OpenAI)}
\end{subfigure}
\begin{subfigure}{0.45\columnwidth}
\centering
\deltaevalplot{piqa}
\caption{PIQA}
\end{subfigure}
\begin{subfigure}{0.45\columnwidth}
\centering
\deltaevalplot{wsc}
\caption{Winogrand Schema Challenge}
\end{subfigure}
\begin{subfigure}{0.45\columnwidth}
\centering
\deltaevalplot{winogrande}
\caption{Winogrande}
\end{subfigure}
\begin{subfigure}{0.45\columnwidth}
\centering
\deltaevalplot{arc_easy}
\caption{AI2 Reasoning Challenge --- Easy Set}
\end{subfigure}
\begin{subfigure}{0.45\columnwidth}
\centering
\deltaevalplot{sciq}
\caption{SciQ}
\end{subfigure}
\caption{Zero-shot evaluations of last Pythia checkpoints prior to the second epoch for deduplicated models.}
\end{figure}

\begin{figure}[h]
\centering
\begin{subfigure}{0.45\columnwidth}
\centering
\evaltimeplot{lambada_openai}{acc}{0}{standard}{north west}{2}
\caption{Standard}
\end{subfigure}
\begin{subfigure}{0.45\columnwidth}
\centering
\evaltimeplotepoch{lambada_openai}{acc}{0}{deduped}{north west}{2}
\caption{Deduplicated}
\end{subfigure}
\caption{LAMBADA (OpenAI) over the course of training. Left is the standard Pile, while the right is the deduplicated Pile. The dashed line indicates where the deduplicated Pile began its second epoch.}
\end{figure}

\begin{figure}[h]
\centering
\begin{subfigure}{0.45\columnwidth}
\centering
\evaltimeplot{wsc}{acc}{0}{standard}{north west}{2}
\caption{Standard}
\end{subfigure}
\begin{subfigure}{0.45\columnwidth}
\centering
\evaltimeplotepoch{wsc}{acc}{0}{deduped}{north west}{2}
\caption{Deduplicated}
\end{subfigure}
\caption{Winogrand Schema Challenge over the course of training. Left is the standard Pile, while the right is the deduplicated Pile. The dashed line indicates where the deduplicated Pile began its second epoch.}
\end{figure}

\begin{figure}[h]
\centering
\begin{subfigure}{0.45\columnwidth}
\centering
\evaltimeplot{winogrande}{acc}{0}{standard}{north west}{2}
\caption{Standard}
\end{subfigure}
\begin{subfigure}{0.45\columnwidth}
\centering
\evaltimeplotepoch{winogrande}{acc}{0}{deduped}{north west}{2}
\caption{Deduplicated}
\end{subfigure}
\caption{Winogrande over the course of training. Left is the standard Pile, while the right is the deduplicated Pile. The dashed line indicates where the deduplicated Pile began its second epoch.}
\end{figure}

\begin{figure}[h]
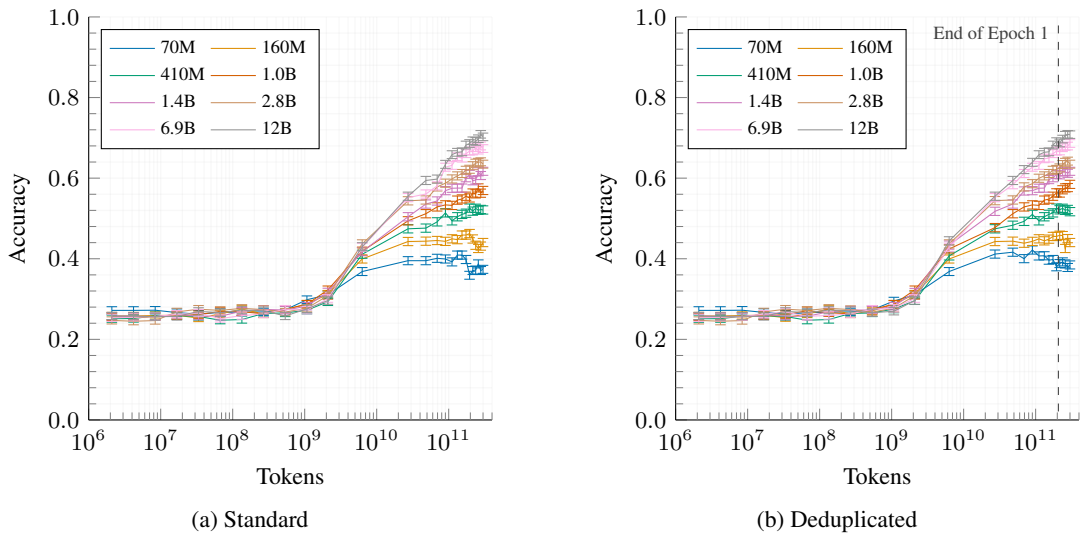

\centering
\begin{subfigure}{0.45\columnwidth}
\centering
\evaltimeplot{arc_easy}{acc}{0}{standard}{north west}{2}
\caption{Standard}
\end{subfigure}
\begin{subfigure}{0.45\columnwidth}
\centering
\evaltimeplotepoch{arc_easy}{acc}{0}{deduped}{north west}{2}
\caption{Deduplicated}
\end{subfigure}
\caption{AI2 Reasoning Challenge --- Easy Set over the course of training. Left is the standard Pile, while the right is the deduplicated Pile. The dashed line indicates where the deduplicated Pile began its second epoch.}
\end{figure}

\begin{figure}[h]
\centering
\begin{subfigure}{0.45\columnwidth}
\centering
\evaltimeplot{sciq}{acc}{0}{standard}{north west}{2}
\caption{Standard}
\end{subfigure}
\begin{subfigure}{0.45\columnwidth}
\centering
\evaltimeplotepoch{sciq}{acc}{0}{deduped}{north west}{2}
\caption{Deduplicated}
\end{subfigure}
\caption{SciQ over the course of training. Left is the standard Pile, while the right is the deduplicated Pile. The dashed line indicates where the deduplicated Pile began its second epoch.}
\end{figure}

\begin{figure}[h]
\centering
\begin{subfigure}{0.45\columnwidth}
\centering
\evaltimeplot{logiqa}{acc}{0}{standard}{north west}{2}
\caption{Standard}
\end{subfigure}
\begin{subfigure}{0.45\columnwidth}
\centering
\evaltimeplotepoch{logiqa}{acc}{0}{deduped}{north west}{2}
\caption{Deduplicated}
\end{subfigure}
\caption{LogiQA over the course of training. Left is the standard Pile, while the right is the deduplicated Pile. The dashed line indicates where the deduplicated Pile began its second epoch.}
\end{figure}

\clearpage

\end{document}